%% file: paper.tex
\documentclass[final]{cvpr}

\input{tex/plots-ext}
\makeatletter
\renewcommand\paragraph{\@startsection{paragraph}{4}{\z@}{1ex}{-1em}{\normalfont\normalsize\bfseries}}
\makeatother

\usepackage{times}
\usepackage{epsfig}
\usepackage{subcaption}
\usepackage{lipsum}
\usepackage{graphicx}
\usepackage{amsmath}
\usepackage{comment}
\usepackage{amssymb}

\usepackage[utf8]{inputenc} 
\usepackage[T1]{fontenc}    
\usepackage{url}            
\usepackage{booktabs}       
\usepackage{amsfonts}       
\usepackage{nicefrac}       
\usepackage{bbm}            
\usepackage{enumitem}
\usepackage{float}
\usepackage{microtype} 
\usepackage[ruled,vlined,linesnumbered]{algorithm2e}

\usepackage[pagebackref=true,breaklinks=true,colorlinks,bookmarks=false]{hyperref}

\begin{document}

\title{A novel shape matching descriptor for real-time hand gesture recognition}

\author{
Michalis Lazarou$^1$ \ \ \ \ Bo Li\ \ \ \ Tania Stathaki$^1$\\
$^1$Imperial College London\\
{\tt\small \{michalis.lazarou14, t.stathaki\}@imperial.ac.uk}
}

\input{tex/abbrev}
\input{tex/defn}
\input{tex/defn-exp}

\maketitle

\begin{abstract}
The current state-of-the-art hand gesture recognition methodologies heavily rely in the use of machine learning. However there are scenarios that machine learning cannot be applied successfully, for example in situations where data is scarce. This is the case when one-to-one matching is required between a query and a dataset of hand gestures where each gesture represents a unique class. In situations where learning algorithms cannot be trained, classic computer vision techniques such as feature extraction can be used to identify similarities between objects. Shape is one of the most important features that can be extracted from images, however the most accurate shape matching algorithms tend to be computationally inefficient for real-time applications. In this work we present a novel shape matching methodology for real-time hand gesture recognition. Extensive experiments were carried out comparing our method with other shape matching methods with respect to accuracy and computational complexity using our own collected hand gesture dataset and a modified version of the MPEG-7 dataset.
Our method outperforms the other methods and provides a good combination of accuracy and computational efficiency for real-time applications. \footnote{Under consideration at Computer Vision and Image Understanding}
\end{abstract}

\input{tex/intro}
\input{tex/related}
\input{tex/method}
\input{tex/experiments}
\input{tex/conclusion}

{\small
\bibliographystyle{ieee_fullname}  
\bibliography{tex/references}  
}
\input{tex/appendix}

\end{document}

%% file: tex/plots-ext.tex
\makeatletter
\@namedef{ver@everyshi.sty}{}
\makeatother

\usepackage[dvipsnames,svgnames,x11names]{xcolor}
\usepackage{tikzextern}
\usepackage{pgffor}

\newcommand{\noextfig}[1]{!!!}
\newcommand{\extdata}[1]{}

%% file: tex/abbrev.tex

\newcommand{\head}[1]{{\smallskip\noindent\textbf{#1}}}
\newcommand{\alert}[1]{{\color{red}{#1}}}
\newcommand{\sm}{\scriptsize}
\newcommand{\eq}[1]{(\ref{eq:#1})}

\newcommand{\Th}[1]{\textsc{#1}}
\newcommand{\mr}[2]{\multirow{#1}{*}{#2}}
\newcommand{\mc}[2]{\multicolumn{#1}{c}{#2}}
\newcommand{\tb}[1]{\textbf{#1}}
\newcommand{\ch}{\checkmark}

\newcommand{\red}[1]{{\color{red}{#1}}}
\newcommand{\blue}[1]{{\color{blue}{#1}}}
\newcommand{\green}[1]{{\color{green}{#1}}}
\newcommand{\gray}[1]{{\color{gray}{#1}}}

\newcommand{\citeme}[1]{\red{[XX]}}
\newcommand{\refme}[1]{\red{(XX)}}

\newcommand{\fig}[2][1]{\includegraphics[width=#1\columnwidth]{fig/#2}}
\newcommand{\figh}[2][1]{\includegraphics[height=#1\columnwidth]{fig/#2}}


\newcommand{\tran}{^\top}
\newcommand{\mtran}{^{-\top}}
\newcommand{\zcol}{\mathbf{0}}
\newcommand{\zrow}{\zcol\tran}

\newcommand{\ind}{\mathbbm{1}}
\newcommand{\expect}{\mathbb{E}}
\newcommand{\nat}{\mathbb{N}}
\newcommand{\zahl}{\mathbb{Z}}
\newcommand{\real}{\mathbb{R}}
\newcommand{\proj}{\mathbb{P}}
\newcommand{\prob}{\mathbf{Pr}}
\newcommand{\normal}{\mathcal{N}}

\newcommand{\mif}{\textrm{if}\ }
\newcommand{\other}{\textrm{otherwise}}
\newcommand{\minimize}{\textrm{minimize}\ }
\newcommand{\maximize}{\textrm{maximize}\ }
\newcommand{\st}{\textrm{subject\ to}\ }

\newcommand{\id}{\operatorname{id}}
\newcommand{\const}{\operatorname{const}}
\newcommand{\sgn}{\operatorname{sgn}}
\newcommand{\var}{\operatorname{Var}}
\newcommand{\mean}{\operatorname{mean}}
\newcommand{\trace}{\operatorname{tr}}
\newcommand{\diag}{\operatorname{diag}}
\newcommand{\vect}{\operatorname{vec}}
\newcommand{\cov}{\operatorname{cov}}
\newcommand{\sign}{\operatorname{sign}}
\newcommand{\prj}{\operatorname{proj}}

\newcommand{\softmax}{\operatorname{softmax}}
\newcommand{\clip}{\operatorname{clip}}

\newcommand{\defn}{\mathrel{:=}}
\newcommand{\peq}{\mathrel{+\!=}}
\newcommand{\meq}{\mathrel{-\!=}}

\newcommand{\floor}[1]{\left\lfloor{#1}\right\rfloor}
\newcommand{\ceil}[1]{\left\lceil{#1}\right\rceil}
\newcommand{\inner}[1]{\left\langle{#1}\right\rangle}
\newcommand{\norm}[1]{\left\|{#1}\right\|}
\newcommand{\abs}[1]{\left|{#1}\right|}
\newcommand{\frob}[1]{\norm{#1}_F}
\newcommand{\card}[1]{\left|{#1}\right|\xspace}
\newcommand{\diff}{\mathrm{d}}
\newcommand{\der}[3][]{\frac{d^{#1}#2}{d#3^{#1}}}
\newcommand{\pder}[3][]{\frac{\partial^{#1}{#2}}{\partial{#3^{#1}}}}
\newcommand{\ipder}[3][]{\partial^{#1}{#2}/\partial{#3^{#1}}}
\newcommand{\dder}[3]{\frac{\partial^2{#1}}{\partial{#2}\partial{#3}}}

\newcommand{\wb}[1]{\overline{#1}}
\newcommand{\wt}[1]{\widetilde{#1}}

\def\xssp{\hspace{1pt}}
\def\ssp{\hspace{3pt}}
\def\msp{\hspace{5pt}}
\def\lsp{\hspace{12pt}}

\newcommand{\cA}{\mathcal{A}}
\newcommand{\cB}{\mathcal{B}}
\newcommand{\cC}{\mathcal{C}}
\newcommand{\cD}{\mathcal{D}}
\newcommand{\cE}{\mathcal{E}}
\newcommand{\cF}{\mathcal{F}}
\newcommand{\cG}{\mathcal{G}}
\newcommand{\cH}{\mathcal{H}}
\newcommand{\cI}{\mathcal{I}}
\newcommand{\cJ}{\mathcal{J}}
\newcommand{\cK}{\mathcal{K}}
\newcommand{\cL}{\mathcal{L}}
\newcommand{\cM}{\mathcal{M}}
\newcommand{\cN}{\mathcal{N}}
\newcommand{\cO}{\mathcal{O}}
\newcommand{\cP}{\mathcal{P}}
\newcommand{\cQ}{\mathcal{Q}}
\newcommand{\cR}{\mathcal{R}}
\newcommand{\cS}{\mathcal{S}}
\newcommand{\cT}{\mathcal{T}}
\newcommand{\cU}{\mathcal{U}}
\newcommand{\cV}{\mathcal{V}}
\newcommand{\cW}{\mathcal{W}}
\newcommand{\cX}{\mathcal{X}}
\newcommand{\cY}{\mathcal{Y}}
\newcommand{\cZ}{\mathcal{Z}}

\newcommand{\vA}{\mathbf{A}}
\newcommand{\vB}{\mathbf{B}}
\newcommand{\vC}{\mathbf{C}}
\newcommand{\vD}{\mathbf{D}}
\newcommand{\vE}{\mathbf{E}}
\newcommand{\vF}{\mathbf{F}}
\newcommand{\vG}{\mathbf{G}}
\newcommand{\vH}{\mathbf{H}}
\newcommand{\vI}{\mathbf{I}}
\newcommand{\vJ}{\mathbf{J}}
\newcommand{\vK}{\mathbf{K}}
\newcommand{\vL}{\mathbf{L}}
\newcommand{\vM}{\mathbf{M}}
\newcommand{\vN}{\mathbf{N}}
\newcommand{\vO}{\mathbf{O}}
\newcommand{\vP}{\mathbf{P}}
\newcommand{\vQ}{\mathbf{Q}}
\newcommand{\vR}{\mathbf{R}}
\newcommand{\vS}{\mathbf{S}}
\newcommand{\vT}{\mathbf{T}}
\newcommand{\vU}{\mathbf{U}}
\newcommand{\vV}{\mathbf{V}}
\newcommand{\vW}{\mathbf{W}}
\newcommand{\vX}{\mathbf{X}}
\newcommand{\vY}{\mathbf{Y}}
\newcommand{\vZ}{\mathbf{Z}}

\newcommand{\va}{\mathbf{a}}
\newcommand{\vb}{\mathbf{b}}
\newcommand{\vc}{\mathbf{c}}
\newcommand{\vd}{\mathbf{d}}
\newcommand{\ve}{\mathbf{e}}
\newcommand{\vf}{\mathbf{f}}
\newcommand{\vg}{\mathbf{g}}
\newcommand{\vh}{\mathbf{h}}
\newcommand{\vi}{\mathbf{i}}
\newcommand{\vj}{\mathbf{j}}
\newcommand{\vk}{\mathbf{k}}
\newcommand{\vl}{\mathbf{l}}
\newcommand{\vm}{\mathbf{m}}
\newcommand{\vn}{\mathbf{n}}
\newcommand{\vo}{\mathbf{o}}
\newcommand{\vp}{\mathbf{p}}
\newcommand{\vq}{\mathbf{q}}
\newcommand{\vr}{\mathbf{r}}
\newcommand{\Vs}{\mathbf{s}}
\newcommand{\vt}{\mathbf{t}}
\newcommand{\vu}{\mathbf{u}}
\newcommand{\vv}{\mathbf{v}}
\newcommand{\vw}{\mathbf{w}}
\newcommand{\vx}{\mathbf{x}}
\newcommand{\vy}{\mathbf{y}}
\newcommand{\vz}{\mathbf{z}}

\newcommand{\vone}{\mathbf{1}}
\newcommand{\vzero}{\mathbf{0}}

\newcommand{\valpha}{{\boldsymbol{\alpha}}}
\newcommand{\vbeta}{{\boldsymbol{\beta}}}
\newcommand{\vgamma}{{\boldsymbol{\gamma}}}
\newcommand{\vdelta}{{\boldsymbol{\delta}}}
\newcommand{\vepsilon}{{\boldsymbol{\epsilon}}}
\newcommand{\vzeta}{{\boldsymbol{\zeta}}}
\newcommand{\veta}{{\boldsymbol{\eta}}}
\newcommand{\vtheta}{{\boldsymbol{\theta}}}
\newcommand{\viota}{{\boldsymbol{\iota}}}
\newcommand{\vkappa}{{\boldsymbol{\kappa}}}
\newcommand{\vlambda}{{\boldsymbol{\lambda}}}
\newcommand{\vmu}{{\boldsymbol{\mu}}}
\newcommand{\vnu}{{\boldsymbol{\nu}}}
\newcommand{\vxi}{{\boldsymbol{\xi}}}
\newcommand{\vomikron}{{\boldsymbol{\omikron}}}
\newcommand{\vpi}{{\boldsymbol{\pi}}}
\newcommand{\vrho}{{\boldsymbol{\rho}}}
\newcommand{\vsigma}{{\boldsymbol{\sigma}}}
\newcommand{\vtau}{{\boldsymbol{\tau}}}
\newcommand{\vupsilon}{{\boldsymbol{\upsilon}}}
\newcommand{\vphi}{{\boldsymbol{\phi}}}
\newcommand{\vchi}{{\boldsymbol{\chi}}}
\newcommand{\vpsi}{{\boldsymbol{\psi}}}
\newcommand{\vomega}{{\boldsymbol{\omega}}}

\newcommand{\rLambda}{\mathrm{\Lambda}}
\newcommand{\rSigma}{\mathrm{\Sigma}}

\newcommand{\vLambda}{\bm{\rLambda}}
\newcommand{\vSigma}{\bm{\rSigma}}

\makeatletter
\newcommand*\bdot{\mathpalette\bdot@{.7}}
\newcommand*\bdot@[2]{\mathbin{\vcenter{\hbox{\scalebox{#2}{$\m@th#1\bullet$}}}}}
\makeatother

\makeatletter
\DeclareRobustCommand\onedot{\futurelet\@let@token\@onedot}
\def\@onedot{\ifx\@let@token.\else.\null\fi\xspace}

\def\eg{\emph{e.g}\onedot} \def\Eg{\emph{E.g}\onedot}
\def\ie{\emph{i.e}\onedot} \def\Ie{\emph{I.e}\onedot}
\def\cf{\emph{cf}\onedot} \def\Cf{\emph{Cf}\onedot}
\def\etc{\emph{etc}\onedot} \def\vs{\emph{vs}\onedot}
\def\wrt{w.r.t\onedot} \def\dof{d.o.f\onedot} \def\aka{a.k.a\onedot}
\def\etal{\emph{et al}\onedot}
\makeatother

%% file: tex/defn.tex
\newcommand{\base}{\mathrm{base}}
\newcommand{\novel}{\mathrm{novel}}
\newcommand{\NN}{\mathrm{NN}}
\newcommand{\masked}{\mathrm{masked}}
\newcommand{\soft}{\mathrm{soft}}

%% file: tex/defn-exp.tex
\newcommand{\ours}{iLPC\xspace}
\newcommand{\lp}{LP\xspace}
\newcommand{\bal}{Balance\xspace}
\newcommand{\ilc}{iLC\xspace}
\newcommand{\cp}{Class\xspace}
\newcommand{\iprob}{iProb\xspace}
\newcommand{\pt}{PT\xspace}

%% file: tex/intro.tex
\section{Introduction}
\label{intro}
Hand gestures are a fundamental aspect in human to human interaction, especially in non-verbal communications \cite{HandSurvey}. Thus it comes as no surprise that research involving hand gestures is growing in the field of human computer interaction. Multiple different applications can benefit through the use of communication with hand gestures such as telerobotics \cite{telerobotics}, virtual reality and augmented reality applications, gaming \cite{gaming} and sign language translation \cite{signlanguage}.

Classic computer vision methods along with machine learning techniques have been proposed over the years in order to facilitate human-computer interaction through the use of extracting semantic information from hand gestures. Some of the main areas of research regarding hand gestures include hand gesture detection, segmentation and recognition \cite{HTS}. In recent years the main line of research that predominated the research community for computer vision problems is the use of machine learning, specifically deep learning \cite{alexNet}.

However, the bottleneck of deep learning is its reliance on vast amount of data, which is not always available in real-world scenarios, making machine learning approaches impossible to train. A representative example could be a scenario where a dataset consists of a large number of classes (i.e. 300) and each class contains only 1 example per class.

In this work we turn our attention to a real-world research problem that can be used in practical applications, real-time hand gesture recognition. Real-time hand gesture recognition has the potential to be used in real-time virtual reality applications, telerobotics and video games. The problem we are addressing can be defined as: given a predefined dataset of $N$ hand gestures and a query hand gesture image, identify the most perceptually similar hand gesture from the dataset. Matching perceptually similar hand gestures can be used as an efficient way to retrieve further information regarding the hand gesture. For example, each hand gesture in the dataset can have additional features associated with it that are calculated offline such as its skeleton information and its 3D shape. Using the matched hand gesture from the dataset, these features can be associated with the query hand gesture as well  and retrieved in real-time. This process eliminates the requirement to calculate these details from scratch and is of great interest in real-time applications that require fast processing, yet they only have limited amount of computational resources, such as mobile phones.

The process of hand gesture matching is part of the complete pipeline of real-time hand gesture recognition following the hand gesture segmentation procedure \cite{rtsegmentation} and alignment procedure where every hand gesture is rotated by its orientation angle, $\theta$\textdegree. Hand gesture segmentation techniques cannot provide 100\% accuracy because a) of the difficulty in segmenting the highly articulated human hands and b) complex backgrounds and varying lighting conditions make the segmentation process challenging \cite{9RThand}. 

In order to address the problem of imperfect segmentation we turn our attention to a class of algorithms that have lost research interest over the years because of deep learning, namely shape matching algorithms. It is hypothesized that a good shape matching algorithm can provide a robust methodology to address shape distortions and other deformations and identify similarities between shapes. Nevertheless, the most accurate shape matching algorithms currently found in literature tend to require high computational resources with non-linear matching complexities. For example the widely used shape matching method shape context has an average time matching complexity of $O(n^3)$ \cite{1SC}.


In this work we propose a novel shape matching algorithm that outperforms some of the most well known shape matching algorithms. Extensive experiments were carried out with respect to two datasets, our collected segmented hand gesture dataset and the MPEG-7 shape matching dataset \cite{mpeg} that is widely used for comparing shape matching algorithms.

To summarize, our contributions are as follows:
\begin{itemize}
    \item We present a novel shape matching problem formulation inspired from a real-world practical application, namely hand gesture matching for real-time applications.
    \item We propose a novel shape matching algorithm that provides a good trade-off with respect to accuracy and computational efficiency for real-time shape matching of highly articulated shapes such as hand gestures.
    \item We propose a novel hand gesture dataset that can be used to compare different shape matching algorithms with respect to accuracy and computational complexity.
\end{itemize}

%% file: tex/related.tex
\section{Background}
\label{background}
\subsection{Hand gesture recognition}
The literature related to hand gesture recognition is divided between dynamic and static hand gesture recognition. Dynamic hand gestures are moving gestures that are represented by a sequence of images while static hand gestures are represented by one image per gesture.  Due to the fact that dynamic gestures require processing of multiple frames they are very challenging for real-time applications and especially for limited computational resources such as mobile phones. Hand gesture recognition approaches can also be separated in techniques that use 3D hand models such as MANO hand model \cite{MANO} or 2D images. Current approaches in 3D hand gesture recognition attempt to regress the 3D model shape and pose parameters \cite{handcvpr, dominikcvpr}. However, 3D methods typically rely on guidance from 2D keypoint detection to regress the model parameters. Therefore, methods that rely on 2D keypoint detection are deemed inappropriate since in our case we only have silhouette images. Furthermore another drawback of most methods attempting to regress the 3D shape is their computational inefficiency. A thorough review regarding 3D hand pose estimation techniques can be found in \cite{3dmotionreview}.

The bulk of 2D static hand gesture methodologies can be classified as either hand-crafted methods or machine learning methods. Hand crafted methods involve methods that extract features from images. A well-known hand-crafted method is the technique of histograms of local orientation used to differentiate between 10 hand gestures \cite{hoghands}. The robustness of local orientation to lighting conditions and the translation invariance achieved by storing the local orientations to a histogram makes this algorithm an interesting approach. Elastic graph matching algorithms have also been used for hand gesture recognition \cite{EGM, HierarchicalEGM}. In \cite{EGM} a hand gesture is viewed as a labeled 2D graph where the nodes of the graph are represented by labelled local image descriptors while the edges of the graph represent distance between the local features in an image. The local features can be any type or a combination of multiple features such as Gabor filters and other texture/shape/colour features. In the matching process, the different graphs of the different hand gestures are applied on the query image and the graph obtaining the highest similarity represents the class of the query.

Deep learning approaches have also been explored recently for 2D static hand gesture recognition \cite{handDL, statichand}. \cite{handDL} showed that biologically inspired neural networks such as convolutional networks and denoising autoencoders can be used successfully for recognition of complex hand gestures.
\cite{RThandsDL} combined deep learning methods with traditional hand-crafted feature extraction methods such as Hu moments and Gabor filters investigating their method on both binary and grayscale images.
\subsection{Shape matching methods for hand gesture recognition}
\label{lit2} 
Most of the hand gesture recognition prior work leverage multiple different features such as colour, texture and shape features. However in real-time settings in order to improve the computational efficiency of the process, binary segmented hand gestures are most commonly used where the only available information is the shape of the gesture. In this section we provide an overview of works that leverage only shape information.

\cite{4RTSC} used the matching with shape context algorithm for real-time hand gesture recognition. In this paper hand gestures are segmented after obtained from a CCD camera. The contours of a segmented hand gesture are sampled, keeping 100 contour points per gesture. The shape context matching method is used to compare the query hand gesture from a dataset of hand gestures. The results were promising since the accuracy of matching the hand gesture under consideration to the most similar hand gesture from the dataset was 70-90\% for the 5 hand gestures used in the paper.

In order to handle noise and the limitations of low-resolution optical sensors such as the Kinect sensor, Zhang Z. and Ren Z. proposed the Finger Earth Mover's distance method to match hand gestures. \cite{FEMD}. This method represents the hand shape as a time series curve in terms of relative distance between each contour point and the centre of the shape against the normalized angle between the starting point on the contour relative to the centre. The method showed excellent results obtaining accuracy of 93.9\% in a designed hand gesture dataset containing 10 hand gestures from different hands.

Fourier descriptors have also been used in various ways for hand gesture recognition purposes. In 2013 Gamal H. used Fourier descriptors for real-time static hand gesture recognition \cite{FDhandGest}. The classic shape signature of representing the 2D shape boundary as a 1D complex function was used as described in \cite{21fourier}. In order to make the shape descriptor scale invariant, N points where selected from the contour of every shape and the Fourier descriptors were obtained using the shape signature.

\cite{handmoments} proposed the idea of representing the hand gesture using Krawtchouk moments, which are derived from Krawtchouk polynomials and are orthogonal. The authors investigated their algorithm on a dataset of 10 classes of hand gestures of 423 examples per class. Comparing Krawtchouk to Zernike and geometric moments, it has been shown experimentally that Krawtchouk moments are more stable, can reconstruct hand gestures better and provide better classification results.

Despite the increasing interest in hand gesture recognition, an open research question that has not been addressed before is how to perform real-time hand gesture recognition when our dataset consists of a large number of classes (more than 100) and each class is only represented from one hand gesture example. None of the aforementioned works addresses this problem but rather they address the problem of having few classes of multiple examples per class \cite{EGM, statichand, handDL, RThandsDL, 4RTSC, FEMD, FDhandGest, 21fourier, handmoments}. In this work we address this novel problem by investigating 2D shape matching techniques.

\subsection{Shape matching methodologies}
Shape matching methods can be classified using different classification methods \cite{13shapesurvey}. The most commonly used way to classify shape matching methods is to divide them between region based and contour based methods. Contour based methods use the shape's boundary points while region based methods use the shape's interior points. In an attempt to provide a coherent overview of the different kinds of shape matching algorithms existing in literature, an artificial classification method is created to provide the backbone for further analysis and discussion. The classification used in this work partitions the shape matching methods into these three categories:
\begin{itemize}
    \item Global shape methods
    \item Transform methods
    \item Point set matching methods
\end{itemize}
\subsubsection{Global shape methods}
There are a lot of methods that work on an object as a whole, either by using the complete area of the shape or by using the boundary of the shape. In order to be used successfully, these methods require the entire object under consideration to be present in the scene in order for its segmentation to be realisable \cite{12SoA}.

\textit{Shape signatures}\\
These methods are 1D dimensional vectors derived from the boundary of the object. It can be thought of as if shape signatures capture the perceptual features of the shape \cite{13shapesurvey}. Some examples of shape signatures include the centroid distance function, area function and the contour curvature. Different shape signatures have different properties and depending on the application of interest, the most appropriate shape signature can be selected. For example, the centroid distance function is expressed as the distance of the boundary points from the centroid of the shape. It is translation invariant since the distance of the contour points to the centre of the shape remains the same regardless how much the shape is translated, yet it is rotation variant since it is sensitive to the starting contour point. On the other hand, contour curvature represents the shape as the curvature of the contour \cite{contourcurvature}. It is translation  and  rotation  invariant  since  the  curvature  of  the  contour  remains the same after translation and rotation. 

\textit{Moments}\\
A shape can also be represented by its shape moments. These can be boundary moments or region moments. Boundary moments can be used to reduce the dimension of the boundary representation \cite{15MomentsReference}. However, region moments are much more common and are more frequently used. In terms of a binary image $f(x,y)$ of dimension $N \times M$, where an object can be clearly identified the raw $(p,q)$-moment of the object is given by:
\begin{center}
   $M_{p,q}= \displaystyle\sum_{x=1}^N\displaystyle\sum_{y=1}^M x^py^qf(x,y)$
\end{center}
where the background pixels are black and have a value of zero and the object pixels are white and have a value of 1. Based on such moments a number of functions such as moment invariants \cite{16Hu} and Zernike moments \cite{17Zernike} can be defined. These functions are invariant under certain deformations such as translation, rotation and scaling \cite{12SoA}.

\textit{Polygonal approximations}
Polygonal approximation methods use the contour of the shape and approximate it with linear piece wise lines. These methods are useful in low resolution images because they reduce the effects of discrete pixelization of the contour. In \cite{polygonal}, Lu C. C. uses polygonal approximation of closed planar shapes as a preprocessing tool in order to eliminate the effect of noise on the shapes.
\subsubsection {Transform methods}
Instead of using shape representations in the 2D space domain, a variety of methods have been proposed to extract shape descriptors in other domains, with the most famous example being the Fourier descriptors. Fourier descriptors are based on the well-known Fourier transform theory \cite{18Fourier}.

\textit{Fourier descriptors}\\
Fourier descriptors are some of the oldest shape processing methods and have been commonly used for over 40 years \cite{13shapesurvey}. Fourier descriptors are usually obtained after Fourier transform is applied on a shape signature or any other shape representation. In 1972 G. H. Granlund was one of the first to propose the use of Fourier descriptors for closed boundary objects for hand print character recognition \cite{21fourier}. He modelled the closed contour, $c(t)$ from the cartesian, $(x,y)$ domain to the complex domain, $c(t) = x(t) + iy(t)$, where $t$ is the contour point index, $t\in\{0,T-1\}$, where the total contour points are $T$. In this way, the contour is represented as a complex 1D sequence. Taking the discrete Fourier transform of the complex function gives the Fourier descriptors:
\begin{center}
   $ a(u) = \frac{1}{N}\displaystyle\sum_{t=0}^{N-1}c(t)\exp{\frac{-i2\pi ut }{N}}, \hspace{5mm}$ for\hspace{1mm} $u=0,1\cdot\cdot\cdot N-1$
    \\\label{FDequ}
\end{center}
A comparative study of several  Fourier descriptors has shown that Fourier descriptors obtained from different shape signatures have different performance levels, with the centroid distance Fourier descriptors outperforming the rest \cite{22FDcomparison}.

\textit{Wavelet descriptors}\\
In 1996 C. H. Chuang and C. C. Kuo proposed a method for boundary representation of shapes based on the wavelet transform \cite{23wavelet}. In this method a hierarchical planar curve descriptor that partitions the boundary into components of different scales is proposed.

\textit{Curvature scale space}\\
In 1999 Mokhtarian F. proposed a shape matching method to address the robustness and computational complexity problems of previous methods by proposing the curvature scale space shape representation \cite{24CSS}. In the proposed method Mokhtarian used a multiscale representation of the curvature of the contour to create a shape descriptor to represent a specific shape. He identified the curvature zero crossing of the contour at different scales, where curvature zero crossings are the points where the sign of the curve changes. At first the curve is represented as a parametric equation:
\begin{equation}
    \Gamma{(u)} = (x(u),y(u))
\end{equation}
In order to create a scale space representation of the contour, each component of the parametric representation of the curve is convolved with a 1D Gaussian Kernel with standard deviation of $\sigma$. The locations of the zero crossings are identified at different scale levels. In the paper the starting scale is $\sigma = 1$ and $\sigma$ gradually increases with $\Delta \sigma = 0.1$. As $\sigma$ increases the number of zero crossings decreases until the curve becomes convex with no zero crossings. The zero crossings at different scales can be represented in the $(u,\sigma)$ plane known as the curvature scale space of the image. Finally the shape representation of this method is the locations of the curvature scale space contour maxima.
\subsubsection {Point set matching methods}
This type of shape representation and matching attempts to identify correspondences between points from two point sets and measures the dissimilarity between them. In contrast to the previous methods where the shape was considered as a whole, the shape is represented as a set of points where each point is considered to be a feature. In some methods these points can be interest points, for example in algorithms such as SIFT and SURF \cite{25SIFT, 26SURF}, in other cases these points can be just some arbitrary sampled contour points, such as in the shape context algorithm \cite{1SC}.

\textit{Hausdorff distance}
Hausdorff distance is a classical correspondence based shape matching algorithm \cite{28}. In 1993, Huttenlocher P. Daniel proposed three different ways of calculating the Hausdorff distance between two binary images efficiently. Given two shapes represented by two point sets, $A = \{a_1, a_2,...,a_p\}$ and $B = \{b_1, b_2,...,b_q\}$ Hausdroff distance is defined as:
\begin{equation}
    H(A,B) = max(h(A,B), h(B,A))
\end{equation}
where
\begin{equation}
    h(A,B) = \displaystyle \max_{a \in A}\min_{b \in B}||a-b||
\end{equation}
Intuitively, Hausdorff distance is the maximum distance of all distances from the points of one point set to their closest point from the other point set.

\textit{Matching with shape context}\\
In 2002, Serge Belongie proposed a novel shape representation and matching algorithm based on finding correspondences between 100 sampled points on the contour of the shape \cite{1SC}. The sampled points are just some arbitrary points chosen at uniform distance from each other. Each sampled point is  described using the shape context descriptor which provides spatial information of where the rest of the sampled contour points are relative to the sampled one.

The Hungarian algorithm is used to find correspondences between the 100 sampled points of two different shapes by solving the assignment problem \cite{hungarian}. Finally the transformation "thin plate spline" is used to align the two shapes together. The dissimilarity between two shapes is proportional to the energy required to align the two shapes and the sum of the Euclidean distance of all the matched points.

\textit{Earth mover's distance}\\
Earth mover's distance (EMD) is a distance metric commonly used in computer vision for matching two weighted point sets \cite{29EMD}. EMD has been used successfully for measuring image similarity in content based image retrieval applications with respect to colour and texture \cite{30EMD, 31EMDretrieval}. It has been claimed that it simulates perceptual distance better than any other distance metric for coloured images \cite{31EMDretrieval}.

Earth mover's distance uses the two point sets provided and treats them as distributions. The two point sets can have different number of points and each point of a point set has a specific weight and a feature vector associated with it. The weight of the point signifies the importance of that specific point in the distribution of the point set. Earth mover's distance calculates the minimum amount of work required to change one point set distribution to the other. This process can be visualized by one distribution being represented as the dirt and the other as the holes. The minimum amount of work to fill the holes with dirt is the earth mover's distance between the two point sets. EMD can be computed by solving an instance of the transportation problem using any algorithm for minimum cost flow such as the Hungarian algorithm \cite{hungarian}.

%% file: tex/method.tex
\section{Methodology}
\label{methodology}
\noindent In this section we will a) describe our problem definition and b) describe our proposed novel shape matching algorithm.
\subsection{Problem specification}
\label{problemdef}
\begin{figure}[H]
    \centering
\begin{tabular}{cc}
    \centering
		\includegraphics[width=0.46\linewidth, height=0.19\textheight]{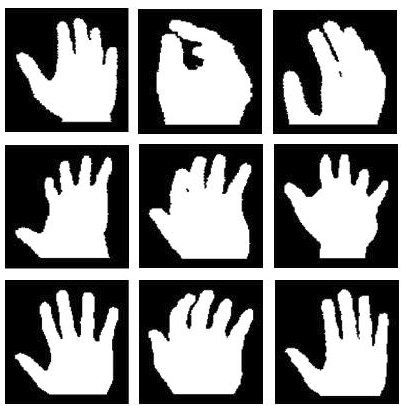}
		&
		\includegraphics[width=0.46\linewidth, height=0.19\textheight]{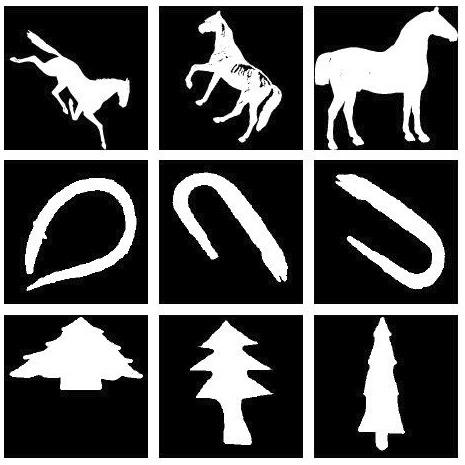} \\
		(a) our dataset & (b) MPEG-7 dataset\\
\end{tabular}
\caption{our dataset compared with MPEG-7 dataset \cite{mpeg}}
\label{fig:datasets}
\end{figure}

It should be noted that this problem is unique and is not aligned with most of the shape matching problems that exist in literature. Some of the fundamental differences of our target problem and the most common shape matching problems addressed in literature are summarized below:
\begin{itemize}
    \item In our dataset, an image represents an individual class while in common shape matching datasets there are multiple images in each class.
    \item Our dataset constitutes of a large number of similar hand gestures where each hand gesture represents a different class. On the other hand, in the most common shape matching problems a much smaller number of classes is used and each class represents a fundamentally different shape than the other class.
    \item in our dataset, very low resolution of images used. This is because we are interested in real-time applications and low resolution of images means lower memory and lower computational power requirements.
\end{itemize}
It can be deduced that the problem we are addressing is harder than the basic shape matching problems found in literature. This is because a) we are trying to distinguish between fine-grained differences between shapes b) all of our shapes are perceptually similar c) a large number of classes is used, d) this problem has not been addressed before in literature which means that there is no prior information on how to address it. In order to visualize the differences between our shape matching problem and the most common shape matching problems addressed in literature, figure \ref{fig:datasets} compares our dataset with one of the most commonly used datasets for 2D shape matching comparisons, the MPEG-7 dataset \cite{mpeg}. Information  regarding  our segmented  hand  gesture  dataset  is summarized in Table \ref{datatable}.

The requirement is to provide an algorithm such that: given a query segmented hand gesture, it can identify the nearest match from our dataset as accurately and efficiently as possible. It should be clarified that in this work we assume that the hand gestures are segmented and aligned before a shape matching algorithm is applied. This is because we are primarily interested in differentiating between very similar yet different hand gestures and complete rotation invariance is not our main goal. Therefore, we are mainly interested robustness to rotation of small degrees to handle inaccuracies from the shape orientation procedure.

\begin{table}[H]

\caption{Hand gesture dataset}
\centering
\begin{tabular}{ |l|l| }
  \hline
  No. of images & $200$ \\
  Type of images & binarized \\
  resolution & $120\times120$ \\
  Images per class & $1$ \\
  Content & hand gestures\\
  \hline
\end{tabular}
\label{datatable}
\end{table}

\subsection{Proposed novel shape matching algorithm}
This is a novel shape matching method. It is inspired from shape context \cite{1SC} and contour points distribution histogram \cite{RadialBins}. The main idea is to capture the shape information of the 2D hand gesture in a global shape descriptor from different 2D rotations of the hand gesture and combine them into one shape descriptor. The intuition behind this idea is that by enclosing information from different views of the hand gesture the descriptor will be more robust to rotation and translation as well as to small degrees of shape deformation. The name of our proposed solution is angular radial bins (ARB).

\subsubsection{Angular radial bins (ARB)}
\label{ARBsection}
Given a binary image of a hand gesture, the first step to calculate the ARB descriptor is to calculate its centroid and its total mass. The total mass is defined as the $0^{th}$-order raw moment given in equation \ref{mass}. The centroid of the shape, $(\bar{x},\bar{y})$,  is calculated using the  first-order  raw moments normalized by the total mass as it is shown by equation \ref{CoM}:
 \begin{equation}
     M_{0,0} = \displaystyle\sum_{x=0}^{N-1}\displaystyle\sum_{y=0}^{N-1} x^0y^0f(x,y)
     \label{mass}
 \end{equation}
 \begin{equation}
    \bar{x} = \frac{M_{1,0}}{M_{0,0}}\hspace{5mm} and\hspace{5mm} \bar{y} = \frac{M_{0,1}}{M_{0,0}}
    \label{CoM}
\end{equation}
In order to make our algorithm more computationally efficient, we will only use information from the contour points of the hand gesture. The contour points are calculated using Suzuki's border following algorithm \cite{suzuki}. In contrast to other algorithms such as \cite{1SC} the contour points are not sampled and all contour points are used. 
In order to make the shape descriptor scale invariant the total mass is used as a scaling factor.

The Euclidean distance between the centroid of the shape and the furthest point on the contour of the shape is used as the radius of the minimum circumscribed circle, $\rho$. The region enclosed from the circumscribed circle is further partitioned into several bins using concentric circles and equal angle intervals. As a starting reference point, 0\textdegree \hspace{0.1mm} is always represented by the vertical line pointing towards the North. Three different variations regarding the value that every bin holds were explored in order to obtain the most accurate histogram based descriptor. The three variations are:
\begin{itemize}
    \item \textbf{Number of contour points}: The number of contour points found in every bin is counted and then divided by the total mass of the shape to make it scale invariant.
    \item \textbf{Accumulative distance}: The Euclidean distance from the centroid to all the contour points is calculated. For every bin the total distance of the contour points inside that bin is calculated and scaled by the total mass of the shape.
    \item \textbf{Average distance}: Instead of giving the total distance of all the contour points found in a bin, the average distance is calculated by dividing the total distance with the number of contour points present in each bin. 
\end{itemize}

\begin{figure}[H]
    \centering
\begin{tabular}{cc}
    \centering
		\includegraphics[width=0.45\linewidth, height=0.35\linewidth]{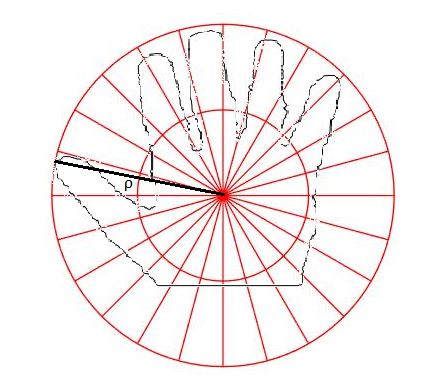}
		&
		\includegraphics[width=0.45\linewidth, height=0.35\linewidth]{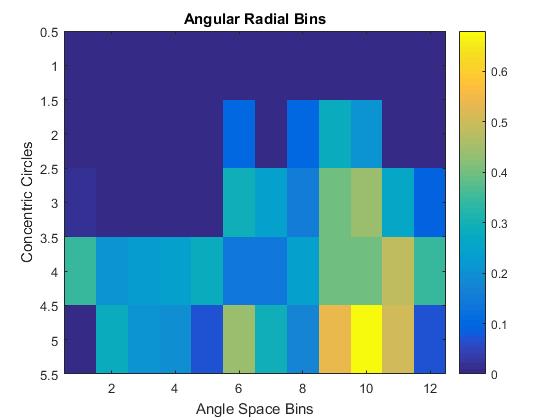} \\
		(a) ARB calculation & (b) Scaled contour points\\
		\includegraphics[width=0.45\linewidth, height=0.35\linewidth]{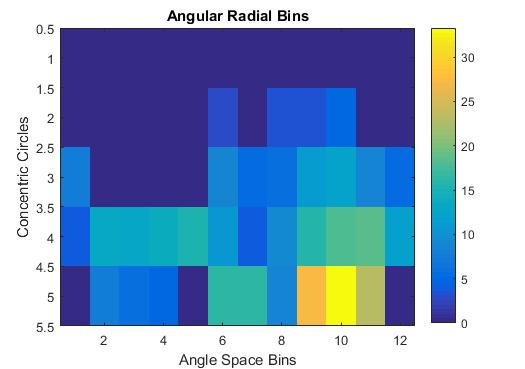} &
		\includegraphics[width=0.45\linewidth, height=0.35\linewidth]{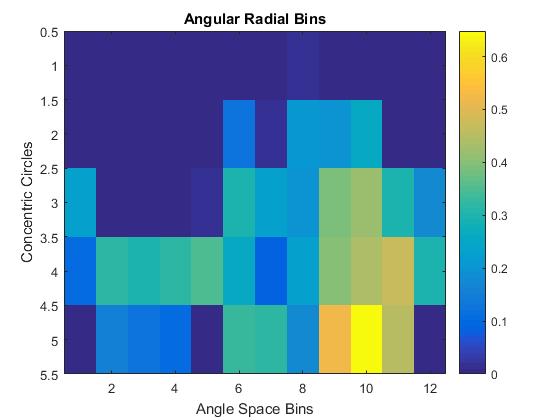}\\
		(c) Accumulative distance & (d) Average distance
\end{tabular}
\caption{ARB descriptor method}
\label{fig:ARB}
\end{figure}

Figure \ref{fig:ARB} shows the calculation procedure and the colour maps of the three different variations using $5$ concentric circles and angular width of $30$\textdegree\hspace{0.1mm}. It can be seen that even though the descriptors look similar there are subtle differences that lead to different results. 

\subsubsection{Overlapping and accumulative angular radial bins}
Hand gestures are extremely articulated shapes which means that even the slightest perturbation such as a slight rotation or even a small deformation of the shape can lead to big changes in the contour points found in each bin and therefore can lead to changes in the weights of every bin. In order to make the shape descriptor more robust and tolerant to changes in rotation and deformation, overlapping and accumulative angular radial bins were proposed. It is hypothesized that calculating the shape descriptor of the same shape but from multiple different 2D rotation and combining this information can potentially make the shape descriptor more robust to deformations, especially rotation.

The concept of using overlapping radial bins to create a histogram-based descriptor is novel and has not been proposed before in the literature. The way that this variation of the shape descriptor works is by obtaining the ARB descriptor as before, however multiple instances of the ARB are applied  onto the shape in such a way so that the bins overlap.

\begin{figure}[h]
    \centering
     \includegraphics[scale = 0.35]{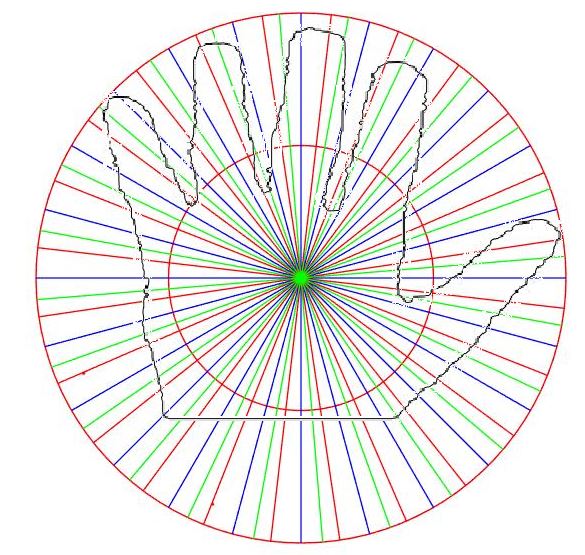}
      \caption{Overlapping angular radial bins}
      \label{fig:overlappingBins}
 \end{figure}

As an example, let the first ARB descriptor with 0\textdegree\hspace{0.1mm} pointing towards the North be exactly as described in Section \ref{ARBsection}. The second ARB descriptor will be added on top of the first one but with a slight angular tilt of $\delta$\textdegree. $N$ different instances of ARB can be added with an accumulative angular tilt of $n\times\delta$\textdegree, where $n$ is the instance number. The relationship between $\theta, \delta$ and $ N $ is defined below:
\begin{center}
    $N = \frac{\theta}{\delta}$
\end{center}
If $N$ is greater than the ratio $\frac{\theta}{\delta}$ then some instances of ARB are already used and are repeated. Figure \ref{fig:overlappingBins} shows the way the shape descriptor is obtained.

\noindent Two different variations were explored in order to obtain the most accurate histogram based descriptor.
\begin{itemize}
    \item \textbf{Overlapping bins}: All the $N$ instances of the ARB shape descriptors are used in order to describe the shape. The descriptor obtained is a 3D histogram in the $\{\rho, \theta, n\}$ space, where $\rho$ is the concentric circle, $\theta$ is the angular space and $n$ represents the instance of the ARB descriptor.
     \item \textbf{Accumulative bins}: The corresponding bins in every instance of ARB are added such that a 2D histogram based descriptor in the $\{\rho, \theta\}$ plane is obtained such that:
     \begin{equation}
         h(\rho_i, \theta_j) = \displaystyle\sum_{n=1}^N h_n(\rho_i, \theta_j)
     \end{equation}
     where: $i$ =  concentric circle $j$ = angular partition, $h_n = $ instance of ARB.
\end{itemize}

\begin{figure}[H]
    \centering
\begin{tabular}{cc}
    \centering
		\includegraphics[width=0.48\linewidth, height=0.35\linewidth]{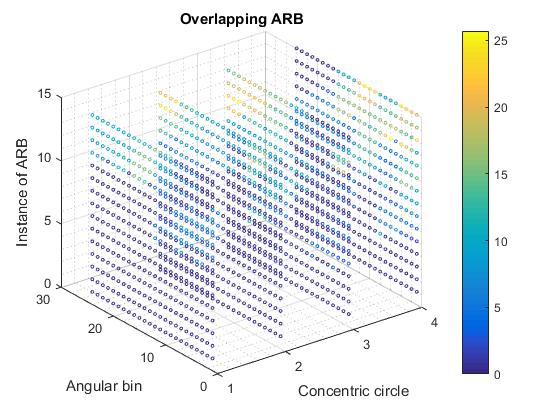}
		&
		\includegraphics[width=0.48\linewidth, height=0.35\linewidth]{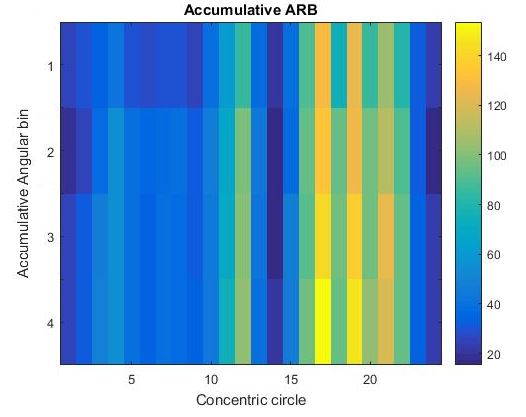} \\
		(a) Overlapping ARB & (b) Accumulative ARB\\
\end{tabular}
\caption{Overlapping and accumulative ARB}
\label{fig:overacc}
\end{figure}

\noindent Figure \ref{fig:overacc} shows the colour maps of the overlapping and accumulative ARB descriptors. Four concentric circles, angular partitioning of 15\textdegree\hspace{0.1mm} leading to $24$ angular bins and $15$ instances of ARB descriptors are used with $\delta = $1\textdegree\hspace{0.1mm} using accumulative distance for the value of every bin.

\subsubsection{Oriented angular radial bins}
\noindent Overlapping/accumulative bins can aid in rotation-tolerance due to holding richer information in how a shape behaves by rotating the ARB descriptor by $\delta$\textdegree. However they cannot provide complete rotation invariance. Three different ways in solving the rotation problem were explored.

A potential way to produce a rotation invariant ARB descriptor would be to use the orientation of the shape, by using the $2^{nd}$ order central moments from equation \ref{centralmoments}:
\begin{equation}
    \mu_{2,0}' = \frac{M_{2,0}}{M_{0,0}} - \bar{x}^2, \mu_{1,1}' = \frac{M_{1,1}}{M_{0,0}} - \bar{x}\bar{y}, \mu_{0,2}' = \frac{M_{0,2}}{M_{0,0}} - \bar{y}^2 
    \label{centralmoments}
\end{equation}
The orientation of the shape is defined by equation \ref{thetamoments} below:
\begin{equation}
    \theta = \frac{1}{2}tan^{-1} \left(\frac{2\mu_{1,1}'}{\mu_{2,0}'-\mu_{0,2}'} \right)
    \label{thetamoments}
\end{equation}

\noindent Using the orientation of the shape the ARB descriptor is shifted so that its starting point, 0\textdegree, is aligned in the direction of the orientation of the shape.
 
Another potential way to solve the orientation problem is to use the 2D DFT on the ARB obtained descriptor. The ARB descriptor is a 2D discretized function in plane $\{\rho,\theta \}$. Let ARB be defined as a 2D discrete function $f(x,y)$, where $x=0,\cdot \cdot \cdot N-1$, and $y=0,\cdot \cdot \cdot M-1$ where $x, y$ represent the concentric circles and the angular space bins. The 2D DFT of $f(x,y)$ function is defined as:
 \begin{equation}
   F(u,v) = \frac{1}{MN} \displaystyle\sum_{x=0}^{N-1}\sum_{y=0}^{M-1}f(x,y)\exp{\Big(-j2\pi\Big( \frac{xu}{N}+\frac{yv}{M}\Big)\Big)}
 \end{equation}
 Using only the magnitude of $F(u,v)$ in a similar way to the way Fourier descriptors are made rotation invariant, the ARB descriptors can be made rotation invariant. This type of rotation invariance was also used from Ying to make the shape context shape matching method rotation invariant \cite{SCFD}.
 
A simpler and rather intuitive way would be to align the 0\textdegree \hspace{0.1mm} of the ARB descriptor along the line made from the centroid of the shape to the contour point that is furthest away in Euclidean distance from the centroid.

Having a histogram based shape descriptor means that simple distance metrics can be used effectively to calculate the distance between 2 hand gestures. The distance between two hand gesture histogram descriptors can be calculated by pointwise subtraction of the two histograms followed by the $\ell_2$-norm of the resulting vector.

%% file: tex/experiments.tex
\section{Experiments}
\label{experiments}
\subsection{Test environment}
The implementations of all the algorithms and all the experiments were implemented in C++ using the openCV library. All the experiments were executed on a workstation Acer Aspire V15, with Intel Core i7-4510U with 2.0GHz speed.
\subsection{Datasets}
\noindent We conduct experiments on two different datasets, our segmented hand gesture dataset and the MPEG-7 dataset that is widely used for comparing shape matching algorithms \cite{mpeg}.

Our own segmented hand gesture dataset consists of 200 unique hand gestures where each hand gesture represents a specific class. Sample images can be seen in section \ref{problemdef} in figure \ref{fig:datasets}.

MPEG-7 dataset consists of 1400 images partitioned into 70 unique classes and each class has 20 image examples. In order to make use of this dataset and transform it to our own requirements, we resized all images to size of $120\times120$ so that the resolution of the images is comparable to our own segmented hand gesture dataset. In order to carry out the one-to-one matching for a large number of classes, 
200 images were randomly sampled and used in the experimental procedure in a similar way to how our hand gesture dataset is utilized.
\subsection{Baseline methods used for comparisons}
In order to get a wider perspective of the performance of our method, we compared it against other shape matching methodologies that have been used for hand gesture recognition.

The algorithms to be investigated were chosen based on their properties and the results published in surveys comparing different methods which have addressed similar problems successfully. For example, curvature scale space and Hu moments were not selected for comparison since, it has been shown in published works, that Fourier descriptors perform better \cite{FDCSS, 11foumom}. 
Matching with shape context \cite{1SC}, finger earth mover's distance \cite{FEMD}, complex coordinates Fourier descriptor \cite{FDhandGest}, centroid distance Fourier descriptor \cite{22FDcomparison} and the Structural feature histogram matrix \cite{2derivatives} were used as baselines.
\subsection{Evaluation protocol}
\subsubsection{Accuracy experiments}
In terms of accuracy evaluation, the requirements that indicate a good shape matching algorithm are tolerance in rotation, translation, scaling, noise and small degrees of deformation of the shape. In order to test the shape matching algorithms based on these criteria, six different deformation tests were designed that test the tolerance of every algorithm for different deformations. The tests can be broken down into three categories as shown in the Table
\ref{deformationTests} below.
\begin{table}[H]
\centering
\caption{Deformation tests}
\label{deformationTests}
\begin{tabular}{|l|l|l|}
\hline
\textbf{Position}         & Translation    & Rotation\\ \hline
\textbf{Point of view} & Scaling     & Perspective     \\ \hline
\textbf{Shape distortion} & Erosion & Dilation \\ \hline
\end{tabular}
\end{table}
The deformation tests work by selecting an image from the dataset as a query image, applying a deformation on the image and then using a shape matching algorithm to find the most similar gestures match from the dataset. 
In order to test the robustness of every algorithm for different levels of deformation, for every one of the six deformation tests, multiple levels of deformation were tested. In the case of our hand gesture dataset 5 different levels of deformation were used while for the MPEG-7 dataset 3 levels of deformation were used. The reason for using only 3 levels of deformation in the MPEG-7 dataset is that some of the shapes that are long and thin, such as snake and spoon shapes, can disappear when high levels of erosion are applied. The $1^{st}$ level of deformation represents the least amount of deformation applied to the query image, while the $5^{th}$ level represents the greatest amount of deformation applied to the query image. For further information and visualization of the tests please refer to the \ref{app:deformations}
\subsubsection{Timing experiments}
The computational efficiencies of all shape matching algorithms are compared through experimental analysis. We are interested in calculating the average training time of an image and the average matching time between two shape representations. The average training time of an image is defined as the time required to obtain a shape representation using the specific shape matching method from a binary image used as an input. The average matching time is defined as the time required to compare two hand gesture shape representations.

In order to calculate the average training and matching time, all the 200 images where used from the dataset. 
The average training time per image is calculated by using 100 iterations of the training procedure for every image.  
The average matching time between two shape representations is estimated in a similar manner, by estimating the average matching time for every possible pair over 10000 iterations.
\subsection{Accuracy results}
\label{accuracysection}
\subsubsection{Method Ablation}
Conceptually the performance of the ARB descriptor can be affected from five different variables. The number of concentric circles, the angular width in degrees of of the angle bins, the assignment of weight of each bin, the number of instances of ARB used in overlapping/accumulative ARB and the different rotation methods used. In this section we break down our method by exploring the effect of each variable separately. We performed our ablation study using our hand gesture dataset. The results and graphs regarding our method ablation can be seen analytically in the appendix \ref{appendixexperiments}.

\subsection{Comparison with other methods}
In this section we compare our method with some of the most well-known shape matching methods applied in hand gesture recognition. We compare all methods using both our segmented hand gesture dataset and the MPEG-7 dataset. In order to provide more details to our results, we provide the percentage of the average matching error for all five different levels of deformation for both datasets in Tables \ref{handall} and \ref{mpegall} and the percentage of the matching error for the $1^{st}$ level of deformation only. The percentage of matching error fir the $1^{st}$ level of deformation provides us with information of the capacity of a method in distinguishing among hand gestures with fine differences only, while the average error provides us with information of how robust each method is, i.e. whether it can still identify the hand gesture when it is heavily distorted.

\begin{table}[H]
\centering
\caption{\% average matching error for every deformation, hand gesture dataset}
\resizebox{\columnwidth}{!}{%
\begin{tabular}{|*{7}{c|}}
 \hline
  \textbf{Algorithm} & \textbf{Trans.}  & \textbf{Scal.} &  \textbf{Ro.} & \textbf{Dil.}  & \textbf{Ero.} &\textbf{Persp.}\\
  \hline
  Shape context \cite{1SC} & 90.8 &60.7  & 85.8 & 89.1 & 73.6 & 93.6 \\
  Finger EMD \cite{FEMD}& \textbf{0} & 25.2 & 54.7 & 68.9& 51.3& 61\\
  CD FD \cite{22FDcomparison} &   \textbf{0} &   15.6 & 17.2 &72.4 &43.5 &27\\
  CC FD \cite{FDhandGest}& 8.2 & 35.4 &34.7 & 60.4&75.4 & 69.9\\
  SFHM \cite{2derivatives} & \textbf{0} & 5.3 & 4.7& 80.8 &62.4 & 33.3\\
  \hline
   \textbf{ARB simple}& \textbf{0} &0.3 & 57.6 & 63.9 &45.6 & 5.2\\
    \textbf{ARB simple or.}& \textbf{0} &1.3 & \textbf{4.6} & 65 &55.1 & 50\\

   \textbf{ARB acc.} &   \textbf{0} &   0.1& 39.5 &62.8&40.9&1.7\\
  \textbf{ARB over.} & \textbf{0} & \textbf{0} & 47.9& \textbf{59.9}& \textbf{39.8} &\textbf{0.9}\\
      \textbf{ARB over. or.} & \textbf{0} & 0.7 & 4.7 &65.4 &54.3 &40.3\\

    \hline
\end{tabular}
}
\label{handall}
\end{table}

\subsection{Timing results}
\label{TimingComp}
The computational efficiencies between the best versions of each of the investigated methods are compared experimentally in Table \ref{timematch}. 
\subsubsection{Dataset image retrieval timing}
To estimate the total time required for each shape matching algorithm in order identify the closest match of a query image from a dataset, the time required to search two datasets of 2000 and 50000 images respectively was estimated through our calculations for the training and matching time from table \ref{timematch} and is summarized in Table \ref{timedatabases}. The images in the datasets are converted to shape descriptors offline. The timing for the matching process consists of: the training time to convert the query image into a shape descriptor, the total matching time needed to match the query shape descriptor to all the shape descriptors of a dataset.

\begin{table}[H]
\centering
\caption{\% matching error for level 1 of every deformation, hand gesture dataset}
\resizebox{\columnwidth}{!}{%
\begin{tabular}{|*{7}{c|}}
 \hline
  \textbf{Algorithm} & \textbf{Trans.}  & \textbf{Scal.} &  \textbf{Ro.} & \textbf{Dil.}  & \textbf{Ero.} &\textbf{Persp.}\\
  \hline
  Shape context \cite{1SC} & 91 & 82.5 & 60 & 79.5 & 65.5 & 92.5 \\
  Finger EMD \cite{FEMD}& \textbf{0} & 14 & 43.5 & 36.5 & 12.5 & 24.8 \\
  CD FD \cite{22FDcomparison} &   \textbf{0}  &   12.5 & 14 & 37.5 &7.5 & 14.3\\
  CC FD \cite{FDhandGest}& 8 & 26.5 & 32.5 &  47.5& 23 & 29.2\\
  SFHM \cite{2derivatives} & \textbf{0} & 1.5 & 5 & 78 & 49.5 & 24.5\\
  \hline
   \textbf{ARB simple} & \textbf{0} & \textbf{0} & 3.5 & 3.5 & 2.5 & 1.8\\
    \textbf{ARB simple or.}& \textbf{0} &0.5 & 1.5 & 8.5 &10 & 7.5\\
   \textbf{ARB acc.} &   \textbf{0} &   \textbf{0} & \textbf{0.5} & 1 & 1 & \textbf{0} \\
  \textbf{ARB over.} & \textbf{0} & \textbf{0} & 1 & \textbf{0} & \textbf{0} & 0.2\\
    \textbf{ARB over. or.} & \textbf{0} & 0.5 & 1.5& 7 & 7.5 &5\\

    \hline
\end{tabular}
}
\label{hand1}
\end{table}

\begin{table}[H]
\centering
\caption{\% average matching error for every deformation,MPEG-7 dataset}
\resizebox{\columnwidth}{!}{%
\begin{tabular}{|*{7}{c|}}
 \hline
  \textbf{Algorithm} & \textbf{Trans.}  & \textbf{Scal.} &  \textbf{Ro.} & \textbf{Dil.}  & \textbf{Ero.} &\textbf{Persp.}\\
  \hline
  Shape context \cite{1SC} & 45.6 & 27 & 41 & 44 & 40.6 & 51.5 \\
  Finger EMD \cite{FEMD}& 0.3 & 29.5 & 43.8 & 44.3 & 37 & 40.2 \\
  CD FD \cite{22FDcomparison} &   \textbf{0} &   30.3 & 16.2 &41.3 & 44.8 & 15.8 \\
  CC FD \cite{FDhandGest}& 14.2 & 46 & 42.1 & 49.8 & 62.8 & 51.3 \\
  SFHM \cite{2derivatives} & 0.5 & 13.8 & 15.8 & 58.3 & 49 & 23\\
  \hline
    \textbf{ARB simple} & \textbf{0} & 4.2 &17.8 &27.3 & 33.2&5.8\\
     \textbf{ARB simple or.} & 0.5 & 9.3 & 19 & 35.2 & 41.7& 20.7\\
   \textbf{ARB acc.} &   \textbf{0} &   3.7 & \textbf{9.3} & 29 & \textbf{29.7}& 6 \\
  \textbf{ARB over.} & \textbf{0} & \textbf{3.3} & 11.8 & \textbf{25}&30.3 &\textbf{4.5}\\
  \textbf{ARB over. or.} & 0.5 & 8.8 & 18.3 & 35.8 & 40.6& 21.7\\

    \hline
\end{tabular}
}
\label{mpegall}
\end{table}

\section{Discussion}
\subsection{Optimization of ARB}
From the experimental in the ablation study in section \ref{accuracysection} we can identify the variables that affect the most the performance of our method. It can be seen from Figure \ref{fig:ARBradial}  that translation and scaling invariance can be achieved with great success by using image moments in order to calculate the centre of mass and the total mass of the shape. In Figure \ref{fig:ARBradial} regarding the concentric circles optimization, it can be seen that the more concentric circles used, the more rotation-tolerant the ARB descriptor becomes. This can be attributed to the fact that the more finely the angular width is partitioned, the bins can capture finer details of the hand gesture shape. A rather surprising result seen in the concentric circles optimization is the inversely proportional relationship between dilation and erosion accuracy performance. It was observed that the more concentric circles used, the more robust the descriptor performed on dilation test while the more concentric circles used the descriptor performed worse on the erosion test.

\begin{table}[H]
\centering
\caption{\% matching error for level 1 of every deformation, MPEG-7 dataset}
\resizebox{\columnwidth}{!}{%
\begin{tabular}{|*{7}{c|}}
 \hline
  \textbf{Algorithm} & \textbf{Trans.}  & \textbf{Scal.} &  \textbf{Ro.} & \textbf{Dil.}  & \textbf{Ero.} &\textbf{Persp.}\\
  \hline
  Shape context \cite{1SC} & 43.5 & 30 & 39 & 29 & 22 & 5 \\
  Finger EMD \cite{FEMD}& \textbf{0} & 26.5 & 42.5& 31& 22 &37 \\
  CD FD \cite{22FDcomparison} &   \textbf{0} &  13 & 15 & 24.5 &17 &14\\
  CC FD \cite{FDhandGest}& 13.5 & 35.5 &39 &40.5 &44 &47.5 \\
  SFHM \cite{2derivatives} & 0.5 & 11.5 & 16.5 & 38.5 & 21.5 &20\\
  \hline
   \textbf{ARB simple} & \textbf{0} & 4 &8.5 &10.5 &10.5 &4.5\\
    \textbf{ARB simple or.} & 0.5 & 10 &13.5 & 17 &18 &14.5\\
   \textbf{ARB acc.} &   \textbf{0} &   4 &\textbf{4} &14 &8.5 & 5\\
  \textbf{ARB over.} & \textbf{0} & \textbf{2.5} & 5.5 & \textbf{10.5} &\textbf{7} &\textbf{3}\\
    \textbf{ARB over or.} & 0.5 & 8 &13.5 & 17.5 &16.5 &15\\
 
    \hline
\end{tabular}
}
\label{mpeg1}
\end{table}

\begin{table}[H]
\centering
\caption{Experimental timing comparisons, hand dataset}
\resizebox{\columnwidth}{!}{%
\begin{tabular}{ |l|l|l|}
  \hline
  \textbf{Algorithm} & \textbf{Matching Time ($\mathbf{\mu}$s)}  & \textbf{Training  time(ms)}\\
  \hline
  Shape context \cite{1SC}& 159400 & 0.2170\\
  Finger EMD \cite{FEMD}& 2.077 &0.6449\\
  Centroid distance FD \cite{22FDcomparison} &   \textbf{0.1856} &   0.1172\\
  Complex co-ordinates FD \cite{FDhandGest} & 0.2139 & \textbf{0.09640}\\
  SFHM \cite{2derivatives} & 0.4928 & 9.4379\\
  \hline
   \textbf{ARB simple} & 0.2193 & 0.1147\\
     \textbf{ARB simple oriented} & 0.2374 & 0.1932\\
    \textbf{ARB accumulative} &   0.2156 &   0.1573\\
  \textbf{ARB overlapping} & 0.6445 & 0.1779\\

    \hline
\end{tabular}
}
\label{timematch}
\end{table}

\begin{table}[H]
\centering
\caption{Total time for hand gesture retrieval from dataset}
\resizebox{\columnwidth}{!}{%
\begin{tabular}{ |l|l|l|}
  \hline
  \textbf{Algorithm} & \textbf{2000 images(ms)}  & \textbf{50000 images(ms)}\\
  \hline
  Shape context \cite{1SC} & 318800 & 7970000\\
  Finger EMD \cite{FEMD}& 4.7989&104.50\\
Centroid distance FD \cite{22FDcomparison} & \textbf{0.4884} & \textbf{9.3972}\\
  Complex co-ordinates FD \cite{FDhandGest} &0.5242 & 10.7914\\
  SFHM \cite{2derivatives} & 10.4235 & 34.0779\\
  \hline
  \textbf{ARB simple} & 0.5533 & 11.0797\\
    \textbf{ARB simple oriented} & 0.668 & 12.0632\\
   \textbf{ARB accumulative} & 0.5885 & 10.9373\\
  \textbf{ARB overlapping} & 1.4675 & 32.4035\\

    \hline
\end{tabular}
}
\label{timedatabases}
\end{table}

It can also be seen from Figure \ref{fig:ARBangular} that smaller angular width improves the rotation performance of the simple ARB. Similarly to the optimization of the concentric circles, the inverse relation between accuracy in dilation and erosion can be observed but in a much lower degree. This leads to the conclusion that the number of concentric circles is more important in providing shape tolerance than the choice of angular width.

In terms of the value stored per bin, it can be seen in Figure \ref{fig:TAP} that the number of contour points and total distance provide the best results with similar performances. Average distance does not perform as well because the number of contour points found per bin should give an indicative value of the importance of that bin in the hand gesture. Therefore average distance cannot provide this information since whether there are 30 contour points found in a bin or 5 their average distance could be exactly the same.

A significant improvement could be seen in Figure \ref{fig:ovacc} in dilation and erosion performance criteria from the use of the overlapping and accumulative ARB descriptor. The overlapping ARB is slightly more accurate than the accumulative ARB because it can hold more information and finer detail. Nevertheless the fact that the accumulative ARB performs in a very similar level of accuracy to that of the overlapping is very positive due to the faster computation time of matching accumulative ARBs.

Using the three methods to improve the rotation problems faced by the ARB descriptor it can be seen from Figure \ref{fig:ARBor} that all three methods are made rotation tolerant with the best rotation solution method being using the $2^{nd}$ order central moments. The significant improvement in rotation tolerance can be seen from the results of the rotation test in section \ref{rotationtests}. Using image moments, the percentage matching error over 200 images remains relatively constant over the five different levels of rotation while the error without image moments keeps increasing at every level of deformation. However using the rotation tolerant methods deteriorates the performance of the shape distortion tests of dilation and erosion and especially perspective.

After the extensive experimental validation in section \ref{accuracysection}, it is demonstrated that the best version of the ARB is the accumulative ARB with 2 concentric circles, 24 angular bins of width 15\textdegree\hspace{1mm} and $\delta$\textdegree\hspace{1mm} angular tilts of 3\textdegree\hspace{1mm} between the different instances of ARB descriptors. The best weight value to assign to the bins is the accumulative distance of all the contour points inside each bin. It should be noted that even though the ARB overlapping performs slightly better than the ARB accumulative, the ARB accumulative provides superior timing performance as it can be seen from Tables \ref{timematch} and \ref{timedatabases} and makes it the most suitable method for real-time hand gesture recognition.


\subsection{Timing results analysis}
It can be seen from Table \ref{timematch} that ARB provides faster training and matching time than all of the implemented shape matching algorithms besides the Fourier descriptors.  The orientation methods used to make the overlapping and accumulative ARB rotation invariant has a very small effect on the training time, while the matching time in theory should be the same due to the same number of bins. Due to its computational efficiency and excellent accuracy performance ARB provides an ideal shape descriptor for real-time shape matching.

From table \ref{timedatabases}, two hypothetical scenarios of image retrieval are shown, retrieving a gesture from 2000 and 50000 image datasets. It can be seen that our proposed solution is only slightly less time efficient than the Fourier descriptors which are the most efficient methods, while at the same time being much more accurate than them. 
\subsection{Comparison with other methods}
Using the hand gesture dataset, it can be seen from Tables \ref{handall} that our algorithm performs better or at least as good as the other algorithm in all the accuracy test criteria. If we take into consideration only the $1^{st}$ level of deformation using table \ref{hand1}, the superiority of our algorithm is even more pronounced, providing the most accurate results by far. In Tables \ref{timematch} and \ref{timedatabases} it can be seen that our algorithm is very computationally efficient, similar in speed to the Fourier descriptor methods but slightly worse. However it is much more accurate than the Fourier descriptor methods, which makes it the best possible choice for hand gesture matching for real-time scenarios.

Besides being appropriate for real-time hand gesture recognition, the applicability of our algorithm in different settings can be seen from tables \ref{mpegall}, \ref{mpeg1} where multiple algorithms are compared against the most well known shape matching dataset, the MPEG-7 dataset \cite{mpeg}. It can be seen that our proposed solution outperforms the other methods in this setting as well as showing that it has generic properties and features that can be applied to general shape matching problems not only for hand gesture. It should be noted that hand-crafted methods specifically addressing hand gestures such as the Finger EMD performed badly in this setting highlighting the limitation of hand-crafted methods for hand gesture matching in different shape matching settings. 

%% file: tex/conclusion.tex
\section{Conclusion}
In this work we proposed a novel shape matching method for a real-time hand gesture recognition. We created a unique dataset of 200 black and white hand gesture images as benchmark and compared our method against various state-of-the-art shape matching method in terms of accuracy and computational efficiency. Our experimental results show that our method significantly outperforms the state-of-the-art methods and provides the optimal trade-off in terms of computational efficiency and accuracy for real-time applications. Also we compared our method with a modified version of the widely used MPEG-7 \cite{mpeg} dataset, outperforming the other methods and showing that our method can provide desired properties for general shape matching problems not just hand gestures. Future directions of research include finding ways to combine the merits of ARB with other methods of complementary strengths such as the SFHM. Also fast search methods could be explored for image retrieval purposes, such as locality sensitive hashing \cite{LSH} and k-d-tree algorithms such as best bin first approach \cite{BBF}. We leave these research directions as open questions for our future research explorations.

%% file: tex/appendix.tex
\appendix
\section{Experiments}
\label{appendixexperiments}

\subsection{Method ablation}
\begin{figure*}
\centering
\begin{tabular}{cc}
    \centering
		\includegraphics[width=0.5\linewidth, height=0.25\textheight]{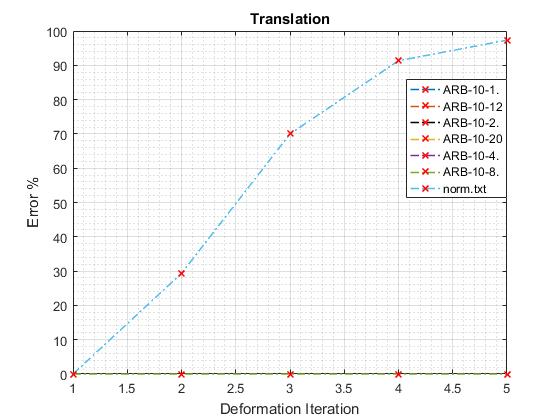}
		&
		\includegraphics[width=0.5\linewidth, height=0.25\textheight]{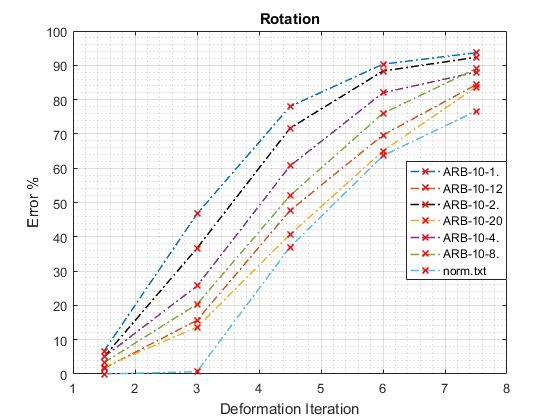} \\
		(a) Translation & (b) Rotation \\
		\includegraphics[width=0.5\linewidth, height=0.25\textheight]{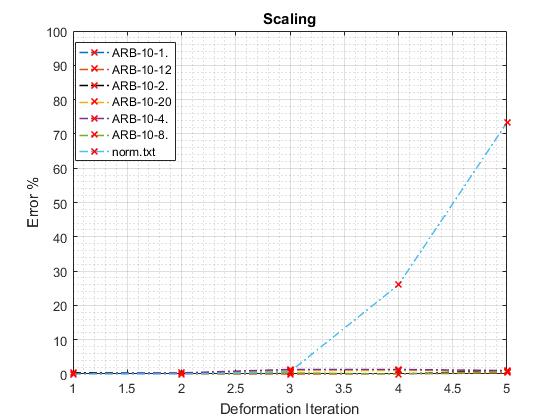} &
		\includegraphics[width=0.5\linewidth, height=0.25\textheight]{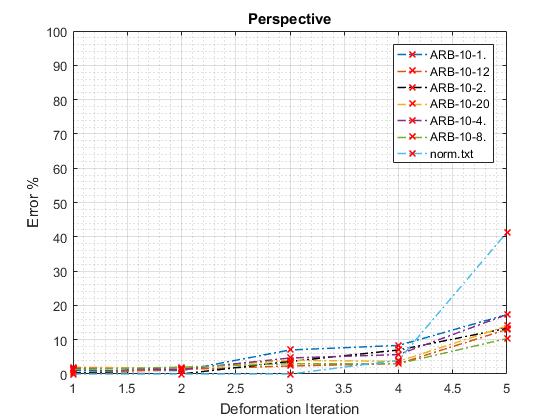}
		\\
		(d) Perspective & (c) Scaling\\
		\includegraphics[width=0.5\linewidth, height=0.25\textheight]{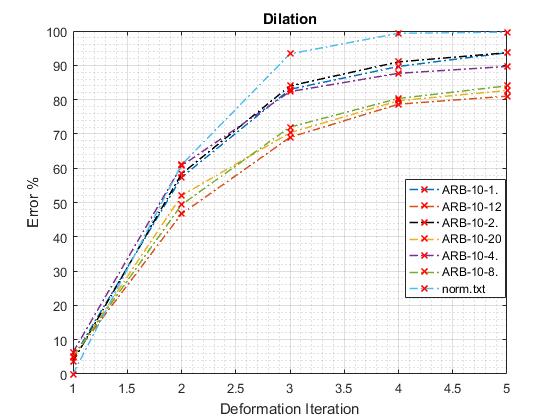} &
		\includegraphics[width=0.5\linewidth, height=0.25\textheight]{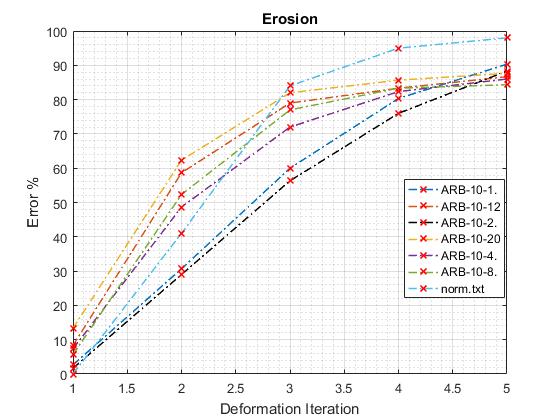}\\
		(e) Dilation & (f) Erosion
\end{tabular}
\caption{ARB concentric circles optimization. \emph{ARB-10-2} represents that the algorithm used is the simple ARB with 10\textdegree\hspace{0.1mm} angular width and 2 concentric circles. \emph{norm} represents the $\ell_2$-norm.}
\label{fig:ARBradial}
\end{figure*}

\noindent \textbf{Optimal quantisation of ARB: number of concentric circles and angular width}\\
In order to identify the best possible number of concentric circles, we compared our method using different number of concentric circles while keeping all the other variables the same. We used a constant width of 10\textdegree\hspace{0.1mm} to partition the angular bins. The weight of each bin represents the total accumulative Euclidean distance from the centroid of the shape to the contour points enclosed in the bin. Figure \ref{fig:ARBradial} shows how the ARB with different concentric circles performed against each other.

\begin{figure*}
\centering
\begin{tabular}{cc}
    \centering
		\includegraphics[width=0.5\linewidth, height=0.25\textheight]{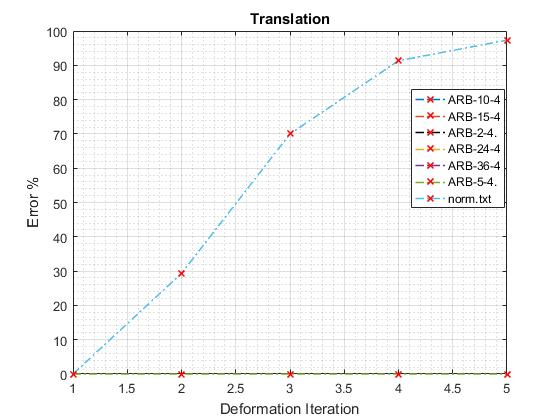}
		&
		\includegraphics[width=0.5\linewidth, height=0.25\textheight]{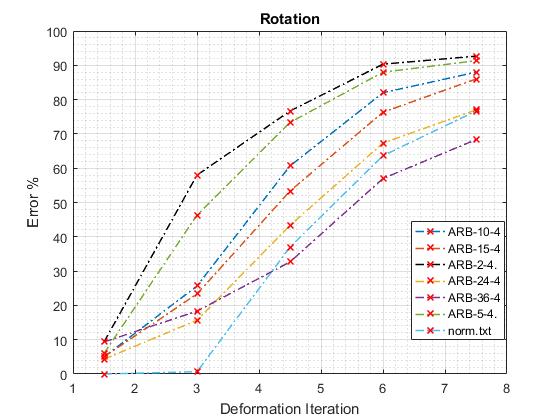} \\
		(a) Translation & (b) Rotation \\
		\includegraphics[width=0.5\linewidth, height=0.25\textheight]{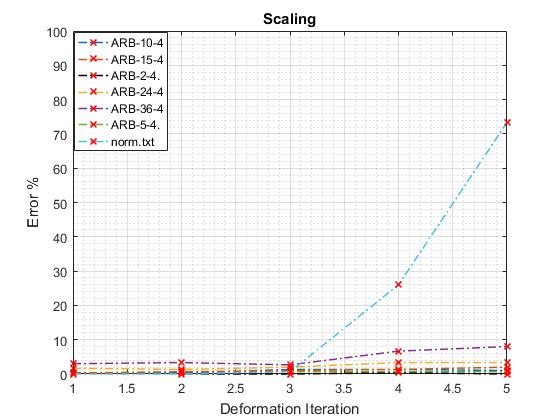} &
		\includegraphics[width=0.5\linewidth, height=0.25\textheight]{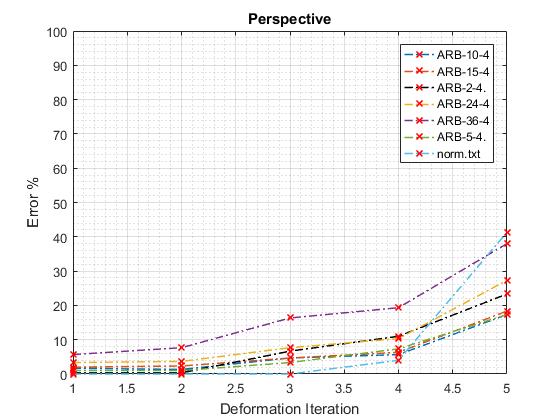}
		\\
		(d) Perspective & (c) Scaling\\
		\includegraphics[width=0.5\linewidth, height=0.25\textheight]{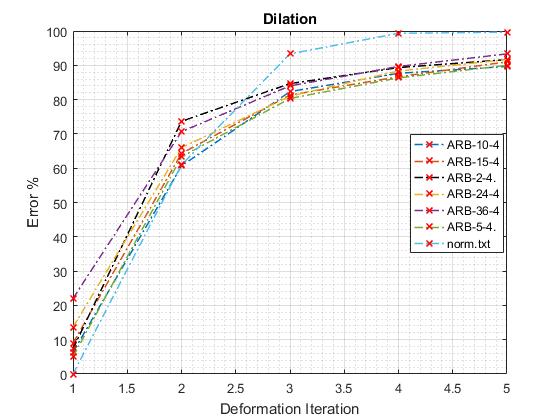} &
		\includegraphics[width=0.5\linewidth, height=0.25\textheight]{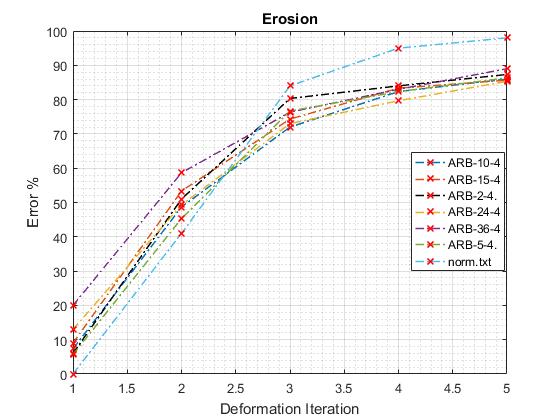}\\
		(e) Dilation & (f) Erosion
\end{tabular}
\caption{ARB angular width optimization. \emph{ARB-10-4} represents that the algorithm used is the simple ARB with 10\textdegree\hspace{0.1mm} angular width and 4 concentric circles. \emph{norm} represents the $\ell_2$-norm.}
\label{fig:ARBangular}
\end{figure*}

In order to identify the best possible angular width in \textdegree\hspace{0.1mm} for each bin, we compared our method using different width sizes in degrees while keeping all the other variables the same. We used a constant number of 4 concentric circles. The weight of each bin represents the total Euclidean distance from the centroid of the shape to the contour points enclosed in the bin. Figure \ref{fig:ARBangular} shows how the ARB with different angular width performed against each other.

\begin{figure*}
\centering
\begin{tabular}{cc}
    \centering
		\includegraphics[width=0.5\linewidth, height=0.25\textheight]{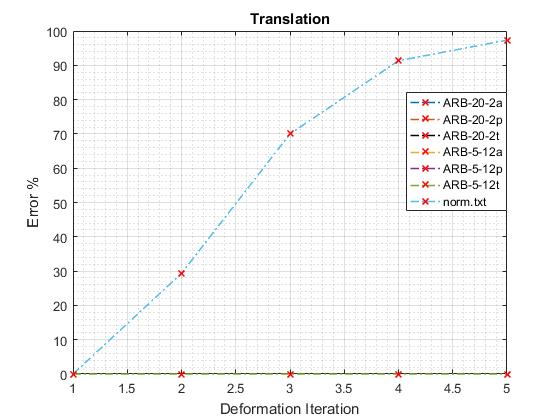}
		&
		\includegraphics[width=0.5\linewidth, height=0.25\textheight]{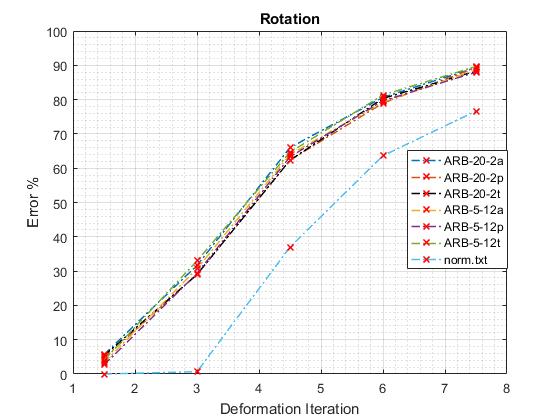} \\
		(a) Translation & (b) Rotation \\
		\includegraphics[width=0.5\linewidth, height=0.25\textheight]{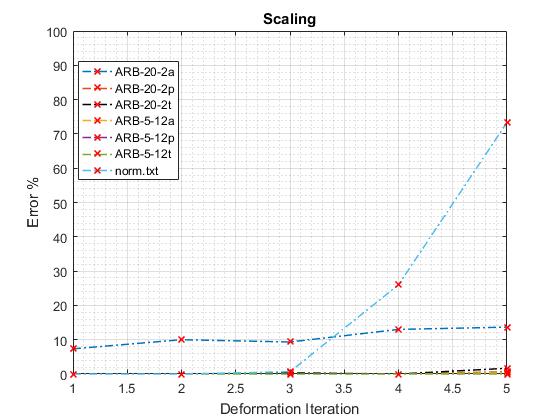} &
		\includegraphics[width=0.5\linewidth, height=0.25\textheight]{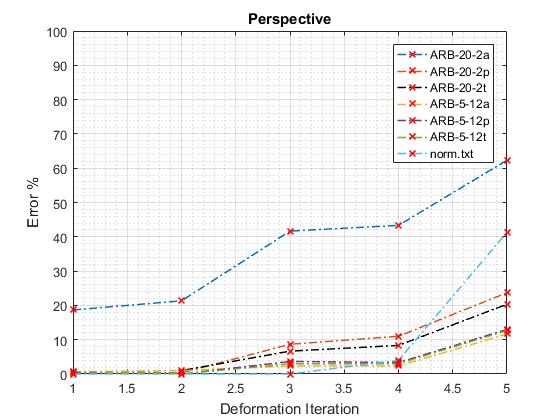}
		\\
		(d) Perspective & (c) Scaling\\
		\includegraphics[width=0.5\linewidth, height=0.25\textheight]{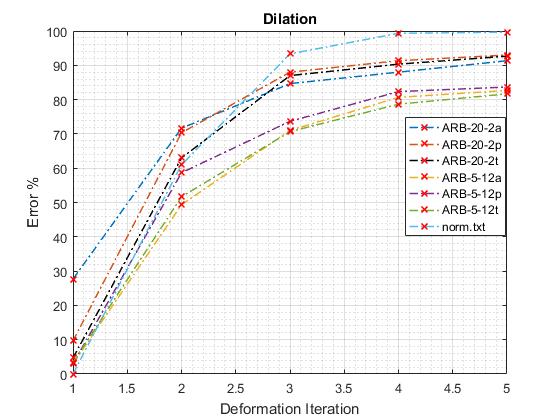} &
		\includegraphics[width=0.5\linewidth, height=0.25\textheight]{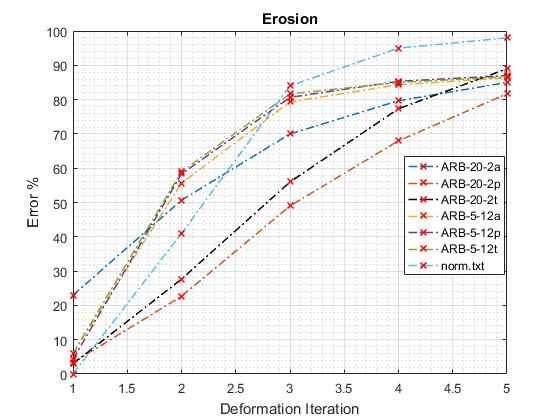}\\
		(e) Dilation & (f) Erosion
\end{tabular}
\caption{Total, average distance, number of contour points per bin compared. \emph{ARB-20-2a} represents that the algorithm used is the simple ARB with 20\textdegree\hspace{0.1mm} angular width and 2 concentric circles with average distance as the weights of the descriptor bins. \emph{ARB-20-2t} represents exactly the same parameters as before apart from using total distance for assigning weights to the descriptor bins. \emph{ARB-10-4p} uses the total number of contour points for assigning weights to the descriptor bins. \emph{norm} represents the $\ell_2$-norm.}
\label{fig:TAP}
\end{figure*}

\noindent \textbf{Comparison of average distance, total distance and total number of contour points}\\
The three different methods for assigning weights to the ARB descriptor bins, namely, calculating the average distance per bin, the accumulative distance per bin and the number of points per bin are compared. Two different ARB descriptors are used, one with 12 concentric circles and 5\textdegree\hspace{1mm} angular width and one with 2 concentric circles and 20\textdegree\hspace{1mm} angular width. The choice of the two configuration was made so that we can investigate the effect of weight assignment for different configurations of ARB with a lot of bins and few bins respectively.  For each ARB descriptor the three ways of giving weight to the ARB descriptor bins are compared. Figure \ref{fig:TAP} shows the results. 

\begin{figure*}
\centering
\begin{tabular}{cc}
    \centering
		\includegraphics[width=0.5\linewidth, height=0.25\textheight]{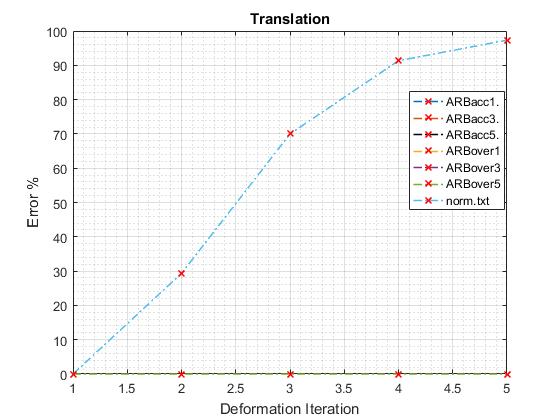}
		&
		\includegraphics[width=0.5\linewidth, height=0.25\textheight]{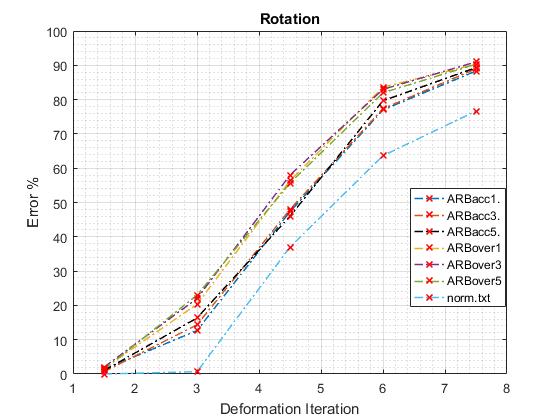} \\
		(a) Translation & (b) Rotation \\
		\includegraphics[width=0.5\linewidth, height=0.25\textheight]{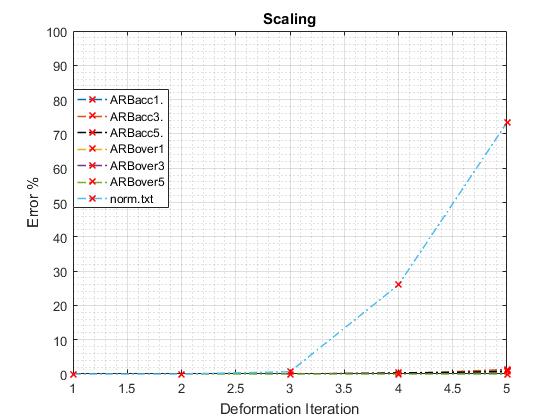} &
		\includegraphics[width=0.5\linewidth, height=0.25\textheight]{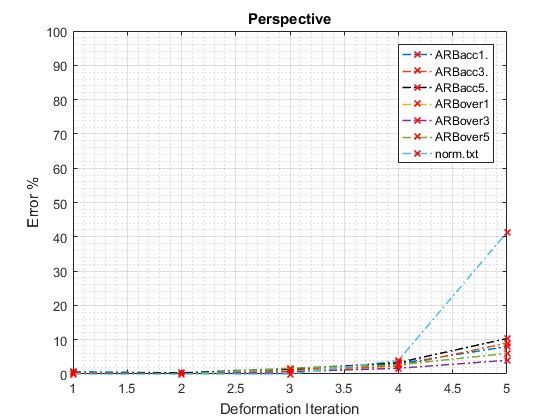}
		\\
		(d) Perspective & (c) Scaling\\
		\includegraphics[width=0.5\linewidth, height=0.25\textheight]{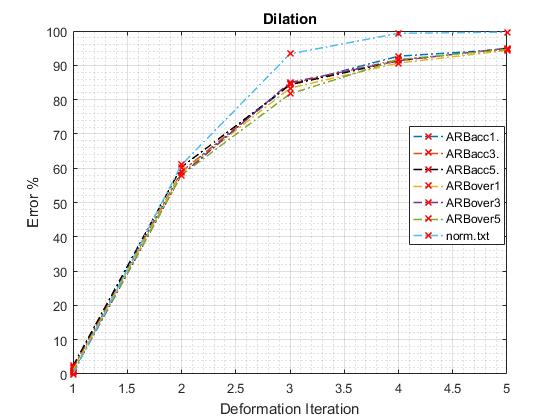} &
		\includegraphics[width=0.5\linewidth, height=0.25\textheight]{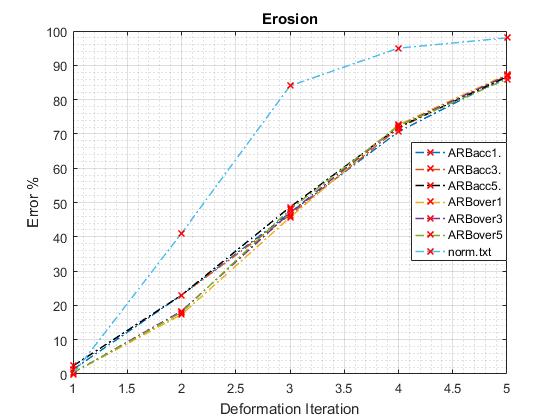}\\
		(e) Dilation & (f) Erosion
\end{tabular}
\caption{Comparison of Overlapping and accumulative angular radial bins. \emph{ARBacc1} represents that the ARB accumulative method is used with $\delta$\textdegree\hspace{0.1mm} tilt=1. \emph{ARBover5} represents that the ARB overlapping method is used with $\delta$\textdegree\hspace{0.1mm} tilt=5.\emph{norm} represents the $\ell_2$-norm.}
\label{fig:ovacc}
\end{figure*}

\noindent \textbf{Overlapping and accumulative ARB}\\
Overlapping and accumulative ARB are compared using the simple ARB of 15\textdegree\hspace{1mm} angular bin size and 2 concentric circles as the basis. Three different $\delta$\textdegree\hspace{1mm} tilts = $\{$1\textdegree, 3\textdegree, 5\textdegree$\}$ are used in order to compare the overlapping and accumulative ARB. 
Figure \ref{fig:ovacc} shows the results.

\begin{figure*}
\centering
\begin{tabular}{cc}
    \centering
		\includegraphics[width=0.5\linewidth, height=0.25\textheight]{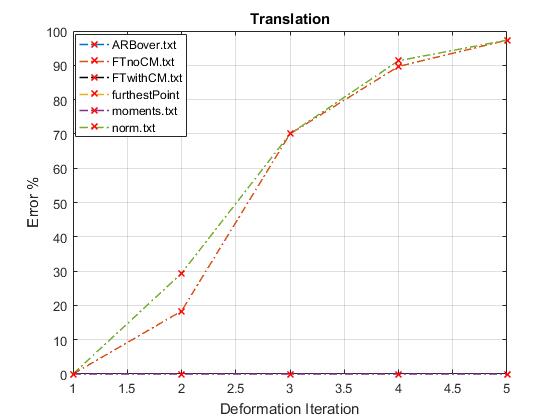}
		&
		\includegraphics[width=0.5\linewidth, height=0.25\textheight]{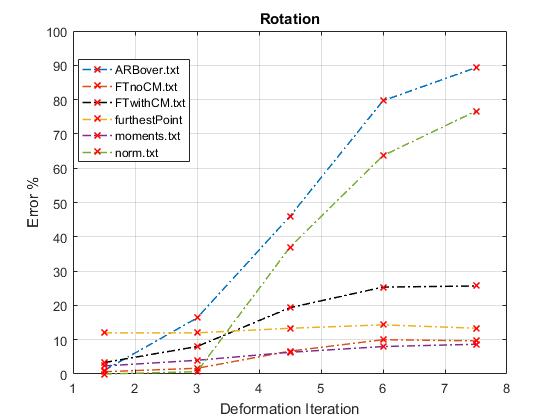} \\
		(a) Translation & (b) Rotation \\
		\includegraphics[width=0.5\linewidth, height=0.25\textheight]{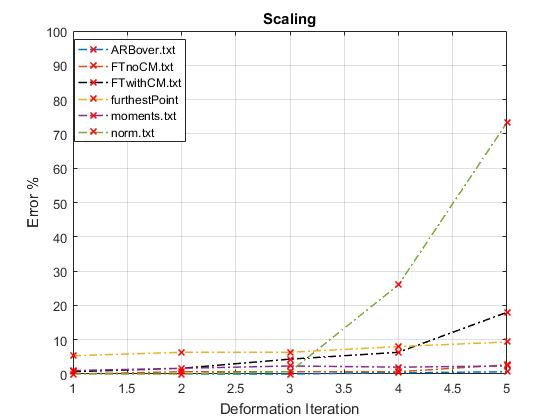} &
		\includegraphics[width=0.5\linewidth, height=0.25\textheight]{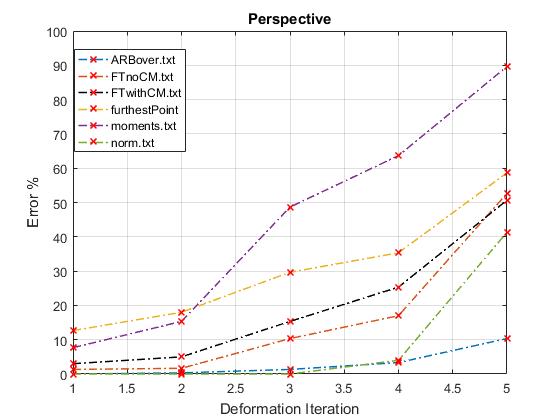}
		\\
		(d) Perspective & (c) Scaling\\
		\includegraphics[width=0.5\linewidth, height=0.25\textheight]{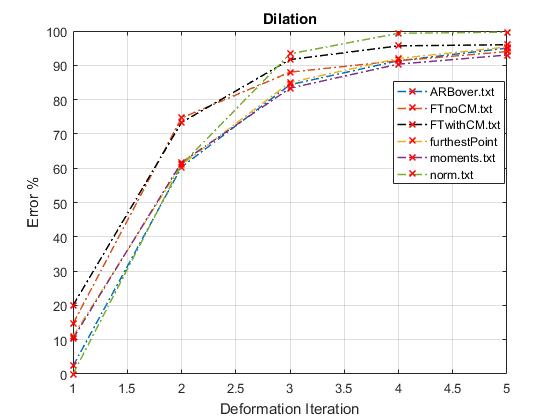} &
		\includegraphics[width=0.5\linewidth, height=0.25\textheight]{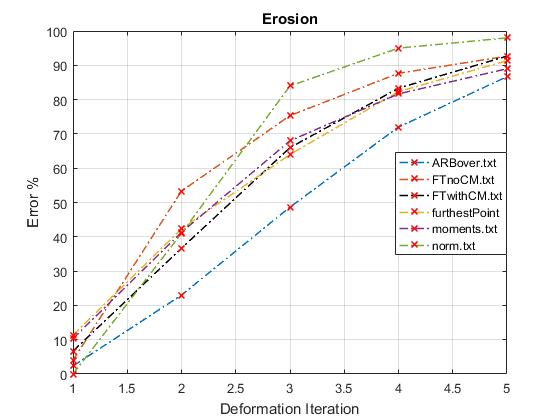}\\
		(e) Dilation & (f) Erosion
\end{tabular}
\caption{Comparison of Rotation tolerance ARB. \emph{ARBover}: ARB overlapping, \emph{FTnoCM}: Fourier Transform without the use of centre of mass, \emph{FTnoCM}: Fourier Transform with the use of centre of mass, \emph{furthestPoit}: furthest point method, \emph{moments}: image moments method. \emph{norm} represents the $\ell_2$-norm.}
\label{fig:ARBor}
\end{figure*}

\noindent \textbf{Rotation Methods Comparison}
\label{rotationtests}
\\Three different methods to make the ARB descriptor rotation tolerant were explored, namely, using Fourier transform, using image moments and making use of the furthest point from the centre of mass of the shape. Fourier transform is also tested using the centre of the image instead of the centre of mass, since it can be made translation invariant by eliminating the DC component. Figure \ref{fig:ARBor} shows the results.

\subsection{Deformation experiments overview}
\label{app:deformations}
\begin{figure*}
	\centering
	\begin{subfigure}{0.7\textwidth}
	    \centering
		\includegraphics[scale=0.3]{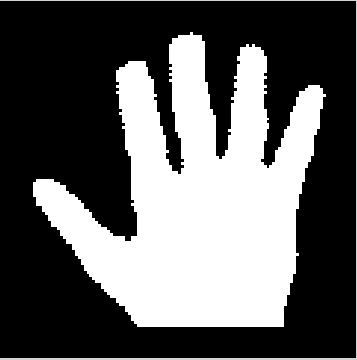}
		\caption{Query image}
	\end{subfigure}
	\begin{subfigure}{0.7\textwidth}
		\centering
		\includegraphics[width=1\linewidth, height =0.095\textheight]{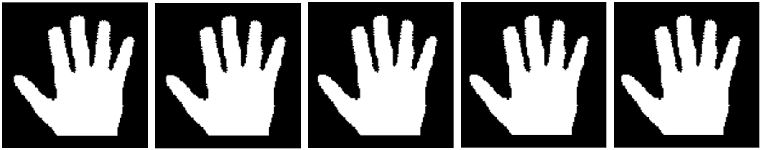}
		\caption{Translation deformation}
	\end{subfigure}
	\begin{subfigure}{0.7\textwidth}
		\centering
		\includegraphics[width=1\linewidth, height =0.095\textheight]{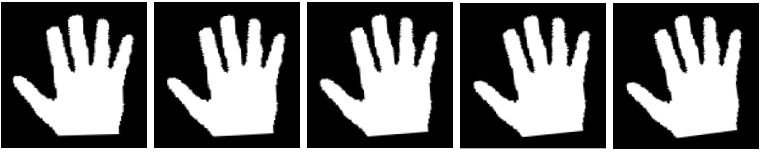}
		\caption{Rotation deformation}
	\end{subfigure}
		\begin{subfigure}{0.7\linewidth}
		\centering
		\includegraphics[width=1\linewidth, height =0.095\textheight]{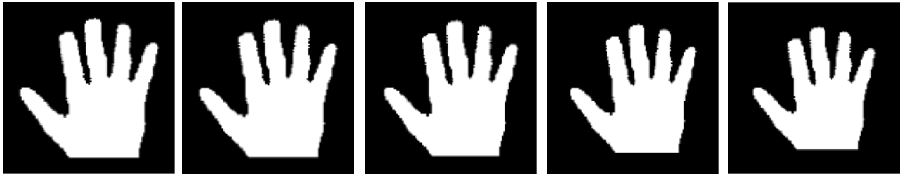}
		\caption{Scaling deformation}
	\end{subfigure}
	\begin{subfigure}{0.7\textwidth}
		\centering
		\includegraphics[width=1\linewidth, height =0.095\textheight]{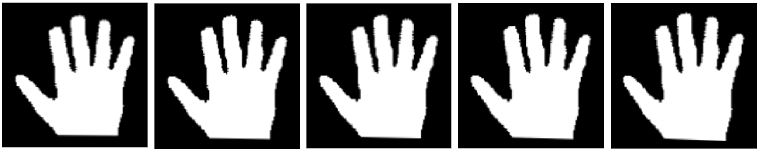}
		\caption{Perspective deformation}
	\end{subfigure}
	\begin{subfigure}{0.7\textwidth}
		\centering
		\includegraphics[width=1\linewidth, height =0.095\textheight]{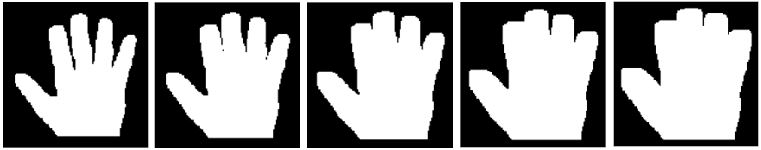}
		\caption{Dilation deformation}
	\end{subfigure}
	\begin{subfigure}{0.7\textwidth}
		\centering
		\includegraphics[width=1\linewidth, height =0.095\textheight]{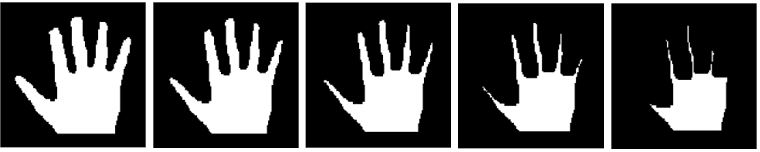}
		\caption{Erosion deformation}
	\end{subfigure}
	\caption{Types of deformations at 5 different levels}
	\label{fig:Transformations}
\end{figure*}

\begin{itemize}
     \item  \textbf{Translation}: The query image is translated in the 4 directions, up, down, left, right. At every level of translation, the query image is translated by 1 pixel in the direction of choice. The average error of the 4 translation directions is used to represent the \% error at that level.
     \item \textbf{Rotation}: The query image is rotated in the two possible directions of the 2D plane, left and right. At every level of rotation, the query image is rotated by 1.5\textdegree \hspace{1mm}in the direction of choice. The average error of the 2 rotation directions is used to represent the \% error at that level.
     \item \textbf{Scaling}: The query image is scaled up and down by a percentage of 1\% of its original size. At every level the average error is calculated for the scaled up and scaled down version.
     \item \textbf{Perspective}: Simulates a 5-level perspective transformation. The 2D image is perceived as if it is a 3D image and is rotated the in the z direction in 5 different levels.
     \item \textbf{Erosion}: The outer layer of pixels around the boundary of the shape is depleted. At every level of erosion, the outer layer of the shape is depleted
     \item \textbf{Dilation}:A new outer layer is added on the shape boundary. This process can be seen as the opposite of erosion. At every level of dilation, an outer layer of 1 pixel thick is added to the shape.
 \end{itemize}
 Figure \ref{fig:Transformations} shows the deformations corresponding to a query image.

%% file: paper.bbl
\begin{thebibliography}{10}\itemsep=-1pt

\bibitem{24CSS}
Sadegh Abbasi, Farzin Mokhtarian, and Josef Kittler.
\newblock Curvature scale space image in shape similarity retrieval.
\newblock {\em Multimedia systems}, 7(6):467--476, 1999.

\bibitem{rtsegmentation}
Reza Azad, Babak Azad, and Iman~Tavakoli Kazerooni.
\newblock Real-time and robust method for hand gesture recognition system based
  on cross-correlation coefficient.
\newblock {\em arXiv preprint arXiv:1408.1759}, 2014.

\bibitem{26SURF}
Herbert Bay, Tinne Tuytelaars, and Luc Van~Gool.
\newblock Surf: Speeded up robust features.
\newblock In {\em European conference on computer vision}, pages 404--417.
  Springer, 2006.

\bibitem{BBF}
Jeffrey~S Beis and David~G Lowe.
\newblock Shape indexing using approximate nearest-neighbour search in
  high-dimensional spaces.
\newblock In {\em Computer Vision and Pattern Recognition, 1997. Proceedings.,
  1997 IEEE Computer Society Conference on}, pages 1000--1006. IEEE, 1997.

\bibitem{1SC}
Serge Belongie, Jitendra Malik, and Jan Puzicha.
\newblock Shape matching and object recognition using shape contexts.
\newblock {\em IEEE transactions on pattern analysis and machine intelligence},
  24(4):509--522, 2002.

\bibitem{handcvpr}
Adnane Boukhayma, Rodrigo~de Bem, and Philip~H.S. Torr.
\newblock 3d hand shape and pose from images in the wild.
\newblock In {\em The IEEE Conference on Computer Vision and Pattern
  Recognition (CVPR)}, June 2019.

\bibitem{29EMD}
Sergio Cabello, Panos Giannopoulos, Christian Knauer, and G{\"u}nter Rote.
\newblock Matching point sets with respect to the earth mover's distance.
\newblock {\em Computational Geometry}, 39(2):118--133, 2008.

\bibitem{17Zernike}
M.~Emre Celebi and Y.~Alp Aslandogan.
\newblock A comparative study of three moment-based shape descriptors.
\newblock In {\em Information Technology: Coding and Computing, 2005. ITCC
  2005. International Conference on}, volume~1, pages 788--793. IEEE, 2005.

\bibitem{RThandsDL}
S{\'e}rgio~F Chevtchenko, Rafaella~F Vale, Valmir Macario, and Filipe~R
  Cordeiro.
\newblock A convolutional neural network with feature fusion for real-time hand
  posture recognition.
\newblock {\em Applied Soft Computing}, 73:748--766, 2018.

\bibitem{23wavelet}
GC-H. Chuang and C-CJ. Kuo.
\newblock Wavelet descriptor of planar curves: Theory and applications.
\newblock {\em IEEE Transactions on Image Processing}, 5(1):56--70, 1996.

\bibitem{30EMD}
S. Cohen.
\newblock {\em Finding Color and Shape Patterns in Images}.
\newblock PhD thesis, Stanford University, Department of Computer Science,
  1999.

\bibitem{11foumom}
Simon Conseil, Salah Bourennane, and Lionel Martin.
\newblock Comparison of fourier descriptors and hu moments for hand posture
  recognition.
\newblock In {\em Signal Processing Conference, 2007 15th European}, pages
  1960--1964. IEEE, 2007.

\bibitem{4RTSC}
Lawrence~Y. Deng, Jason~C. Hung, Huan-Chao Keh, Kun-Yi Lin, Yi-Jen Liu,
  Nan-Ching Huang, et~al.
\newblock Real-time hand gesture recognition by shape context based matching
  and cost matrix.
\newblock {\em JNW}, 6(5):697--704, 2011.

\bibitem{3dmotionreview}
Ali Erol, George Bebis, Mircea Nicolescu, Richard~D Boyle, and Xander Twombly.
\newblock Vision-based hand pose estimation: A review.
\newblock {\em Computer Vision and Image Understanding}, 108(1-2):52--73, 2007.

\bibitem{hoghands}
William~T Freeman and Michal Roth.
\newblock Orientation histograms for hand gesture recognition.
\newblock In {\em International workshop on automatic face and gesture
  recognition}, volume~12, pages 296--301, 1995.

\bibitem{FDhandGest}
Heba~M. Gamal, HM. Abdul-Kader, and Elsayed~A. Sallam.
\newblock Hand gesture recognition using fourier descriptors.
\newblock In {\em Computer Engineering \& Systems (ICCES), 2013 8th
  International Conference on}, pages 274--279. IEEE, 2013.

\bibitem{21fourier}
G{\"o}sta~H Granlund.
\newblock Fourier preprocessing for hand print character recognition.
\newblock {\em IEEE transactions on computers}, 100(2):195--201, 1972.

\bibitem{16Hu}
Ming-Kuei Hu.
\newblock Visual pattern recognition by moment invariants.
\newblock {\em IRE transactions on information theory}, 8(2):179--187, 1962.

\bibitem{HTS}
Noor~Adnan Ibraheem and Rafiqul~Zaman Khan.
\newblock Survey on various gesture recognition technologies and techniques.
\newblock {\em International journal of computer applications}, 50(7), 2012.

\bibitem{alexNet}
Alex Krizhevsky, Ilya Sutskever, and Geoffrey~E. Hinton.
\newblock Imagenet classification with deep convolutional neural networks.
\newblock In {\em Advances in neural information processing systems}, pages
  1097--1105, 2012.

\bibitem{hungarian}
Harold~W. Kuhn.
\newblock The hungarian method for the assignment problem.
\newblock {\em Naval Research Logistics (NRL)}, 2(1-2):83--97, 1955.

\bibitem{dominikcvpr}
Dominik Kulon, Riza~Alp G{\"u}ler, Iasonas Kokkinos, Michael Bronstein, and
  Stefanos Zafeiriou.
\newblock Weakly-supervised mesh-convolutional hand reconstruction in the wild.
\newblock {\em arXiv preprint arXiv:2004.01946}, 2020.

\bibitem{HierarchicalEGM}
Yu-Ting Li and Juan~P Wachs.
\newblock Hierarchical elastic graph matching for hand gesture recognition.
\newblock In {\em Iberoamerican Congress on Pattern Recognition}, pages
  308--315. Springer, 2012.

\bibitem{25SIFT}
David~G. Lowe.
\newblock Distinctive image features from scale-invariant keypoints.
\newblock {\em International journal of computer vision}, 60(2):91--110, 2004.

\bibitem{polygonal}
Cheng-Chang Lu and James~George Dunham.
\newblock Shape matching using polygon approximation and dynamic alignment.
\newblock {\em Pattern Recognition Letters}, 14(12):945--949, 1993.

\bibitem{handDL}
Oyebade~K Oyedotun and Adnan Khashman.
\newblock Deep learning in vision-based static hand gesture recognition.
\newblock {\em Neural Computing and Applications}, 28(12):3941--3951, 2017.

\bibitem{contourcurvature}
Fred Park.
\newblock Shape descriptor/feature extraction techniques.
\newblock {\em UCI iCAMP2011}, pages 1--25, 2011.

\bibitem{signlanguage}
Becky~Sue Parton.
\newblock Sign language recognition and translation: A multidisciplined
  approach from the field of artificial intelligence.
\newblock {\em Journal of deaf studies and deaf education}, 11(1):94--101,
  2006.

\bibitem{18Fourier}
Eric Persoon and King-Sun Fu.
\newblock Shape discrimination using fourier descriptors.
\newblock {\em IEEE Transactions on systems, man, and cybernetics},
  7(3):170--179, 1977.

\bibitem{statichand}
Raimundo~F Pinto, Carlos~DB Borges, Ant{\^o}nio Almeida, and I{\'a}lis~C Paula.
\newblock Static hand gesture recognition based on convolutional neural
  networks.
\newblock {\em Journal of Electrical and Computer Engineering}, 2019, 2019.

\bibitem{handmoments}
S~Padam Priyal and Prabin~Kumar Bora.
\newblock A robust static hand gesture recognition system using geometry based
  normalizations and krawtchouk moments.
\newblock {\em Pattern Recognition}, 46(8):2202--2219, 2013.

\bibitem{mpeg}
Richard Ralph.
\newblock Mpeg-7 core experiment ce-shape-1.
\newblock 1999.

\bibitem{HandSurvey}
Siddharth~S. Rautaray and Anupam Agrawal.
\newblock Vision based hand gesture recognition for human computer interaction:
  a survey.
\newblock {\em Artificial Intelligence Review}, 43(1):1--54, 2015.

\bibitem{FEMD}
Zhou Ren, Junsong Yuan, and Zhengyou Zhang.
\newblock Robust hand gesture recognition based on finger-earth mover's
  distance with a commodity depth camera.
\newblock In {\em Proceedings of the 19th ACM international conference on
  Multimedia}, pages 1093--1096. ACM, 2011.

\bibitem{MANO}
Javier Romero, Dimitrios Tzionas, and Michael~J Black.
\newblock Embodied hands: Modeling and capturing hands and bodies together.
\newblock {\em ACM Transactions on Graphics (ToG)}, 36(6):245, 2017.

\bibitem{31EMDretrieval}
Yossi Rubner, Carlo Tomasi, and Leonidas~J. Guibas.
\newblock The earth mover's distance as a metric for image retrieval.
\newblock {\em International journal of computer vision}, 40(2):99--121, 2000.

\bibitem{RadialBins}
Xin Shu and Xiao-Jun Wu.
\newblock A novel contour descriptor for 2d shape matching and its application
  to image retrieval.
\newblock {\em Image and vision Computing}, 29(4):286--294, 2011.

\bibitem{15MomentsReference}
Milan Sonka, Vaclav Hlavac, and Roger Boyle.
\newblock Image processing, analysis and machine vision.
\newblock {\em Chapman \& Hall Computer series}, pages 259--262, 1993.

\bibitem{9RThand}
Ekaterini Stergiopoulou, Kyriakos Sgouropoulos, Nikos Nikolaou, Nikos
  Papamarkos, and Nikos Mitianoudis.
\newblock Real time hand detection in a complex background.
\newblock {\em Engineering Applications of Artificial Intelligence}, 35:54--70,
  2014.

\bibitem{suzuki}
Satoshi Suzuki et~al.
\newblock Topological structural analysis of digitized binary images by border
  following.
\newblock {\em Computer vision, graphics, and image processing}, 30(1):32--46,
  1985.

\bibitem{EGM}
Jochen Triesch and Christoph Von~Der Malsburg.
\newblock A system for person-independent hand posture recognition against
  complex backgrounds, 2001.

\bibitem{12SoA}
Remco~C Veltkamp and Michiel Hagedoorn.
\newblock State of the art in shape matching.
\newblock In {\em Principles of visual information retrieval}, pages 87--119.
  Springer, 2001.

\bibitem{LSH}
Jingdong Wang, Heng~Tao Shen, Jingkuan Song, and Jianqiu Ji.
\newblock Hashing for similarity search: A survey.
\newblock {\em arXiv preprint arXiv:1408.2927}, 2014.

\bibitem{13shapesurvey}
Mingqiang Yang, Kidiyo Kpalma, and Joseph Ronsin.
\newblock A survey of shape feature extraction techniques.
\newblock In-Tech, 2008.

\bibitem{SCFD}
Su Yang and Yuanyuan Wang.
\newblock Rotation invariant shape contexts based on feature-space fourier
  transformation.
\newblock In {\em Image and Graphics, 2007. ICIG 2007. Fourth International
  Conference on}, pages 575--579. IEEE, 2007.

\bibitem{FDCSS}
Dengsheng Zhang and Guojun Lu.
\newblock A comparative study of curvature scale space and fourier descriptors
  for shape-based image retrieval.
\newblock {\em Journal of Visual Communication and Image Representation},
  14(1):39--57, 2003.

\bibitem{28}
Dengsheng Zhang and Guojun Lu.
\newblock Review of shape representation and description techniques.
\newblock {\em Pattern recognition}, 37(1):1--19, 2004.

\bibitem{22FDcomparison}
Dengsheng Zhang, Guojun Lu, et~al.
\newblock A comparative study of fourier descriptors for shape representation
  and retrieval.
\newblock In {\em Proc. of 5th Asian Conference on Computer Vision (ACCV)},
  pages 646--651. Citeseer, 2002.

\bibitem{2derivatives}
Jing Zhang and Liu Wenyin.
\newblock A pixel-level statistical structural descriptor for shape measure and
  recognition.
\newblock In {\em Document Analysis and Recognition, 2009. ICDAR'09. 10th
  International Conference on}, pages 386--390. IEEE, 2009.

\bibitem{telerobotics}
Hao Zhong, Juan~P Wachs, and Shimon~Y Nof.
\newblock A collaborative telerobotics network framework with hand gesture
  interface and conflict prevention.
\newblock {\em International Journal of Production Research},
  51(15):4443--4463, 2013.

\bibitem{gaming}
Yanmin Zhu and Bo Yuan.
\newblock Real-time hand gesture recognition with kinect for playing racing
  video games.
\newblock In {\em 2014 International Joint Conference on Neural Networks
  (IJCNN)}, pages 3240--3246. IEEE, 2014.

\end{thebibliography}
